\documentclass[10pt,twocolumn,letterpaper]{article}

\usepackage{cvpr}              

\usepackage{bbm}
\usepackage{bm} 
\usepackage{listings}
\usepackage{adjustbox}
\usepackage{booktabs} 
\usepackage{multirow}

\usepackage{pgf}
\usepackage{pgfplots}
\pgfplotsset{compat=1.18}

\usepackage[most]{tcolorbox}
\usepackage{listings}
\usepackage{float}
\usepackage[x11names]{xcolor} 

\usepackage{booktabs}       
\usepackage{amsfonts}       
\usepackage{nicefrac}       
\usepackage{microtype}      
\usepackage{xcolor}         
\usepackage{microtype}
\usepackage{url}
\usepackage{booktabs}
\usepackage{amssymb}
\usepackage[utf8]{inputenc} 
\usepackage[T1]{fontenc}    
\usepackage{url}            
\usepackage{booktabs}       
\usepackage{amsfonts}       
\usepackage{nicefrac}       
\usepackage{microtype}      
\usepackage{xcolor}         
\usepackage{colortbl}
\usepackage{graphicx}
\usepackage{amsmath}
\usepackage{makecell}
\usepackage{multirow}
\usepackage{pifont}
\usepackage[toc,page,header]{appendix}
\usepackage{minitoc}
\usepackage{csquotes}
\usepackage{subcaption}
\usepackage{caption}
\usepackage{adjustbox}
\usepackage{colortbl}
\usepackage[most]{tcolorbox}
\usepackage{lineno}
\usepackage{pgf}
\usepackage{pgfplots}
\usepackage{booktabs}
\usepackage{scalerel}
\usepackage{float}
\usepackage[table,dvipsnames]{xcolor}
\usepackage{booktabs}
\usepackage{caption}
\usepackage{nicematrix} 
\usepackage{tikz} 
\usepackage{xcolor}
\usepackage{soul}             
\usepackage{bm} 
\usepackage{enumitem}
\usepackage{svg}
\usepackage{xspace}
\usepackage{dsfont}

\usepackage{longtable}
\newtcolorbox{planbox}[1]{
    colback=gray!5,
    colframe=gray!75,
    title=#1,
    fonttitle=\bfseries
}

\tcbset{colback=white}
\colorlet{titleblue}{blue!80!black}
\colorlet{titlered}{red!80!black}
\colorlet{titlegreen}{green!80!black}
\colorlet{darkgreen}{green!50!black}

\definecolor{logoRed}{HTML}{CB2F10}
\definecolor{logoBlue}{HTML}{1A4E8A}
\definecolor{logoCyan}{HTML}{56BBCC}

\definecolor{genLLM}{RGB}{250,250,250} 
\definecolor{specLLM}{RGB}{230,230,230}
\definecolor{ourLLM}{RGB}{230,242,230}  
\definecolor{spaLLM}{RGB}{255,249,196}  

\definecolor{genVLM}{RGB}{255,255,255}    
\definecolor{ourVLM}{RGB}{210,210,210}

\tcbset{
  aibox/.style={
    width=\linewidth,
    top=10pt,
    colback=white,
    colframe=black,
    colbacktitle=black,
    enhanced,
    center,
    attach boxed title to top left={yshift=-0.1in,xshift=0.15in},
    boxed title style={boxrule=0pt,colframe=white,},
  }
}

\newtcolorbox{AIbox}[2][]{aibox,title=#2,#1}

\newcommand{\modelnamenc}{{GazeAnywhere}}

\usepackage{tikz}
\makeatletter
\definecolor{applegreen}{rgb}{0.55, 0.71, 0.0}

\newcommand{\ccmark}{\color{applegreen} \ding{51}}%
\newcommand{\cxmark}{\color{red} \ding{55}}%

\makeatother

\definecolor{cvprblue}{rgb}{0.21,0.49,0.74}
\usepackage[pagebackref,breaklinks,colorlinks,allcolors=cvprblue]{hyperref}

\def\paperID{***} 
\def\confName{CVPR}
\def\confYear{2026}

\title{Gaze Target Estimation Anywhere with Concepts}

\author{
Xu Cao$^{1}$, Houze Yang$^{1}$\thanks{Co-first author}, Vipin Gunda$^{1}$, Zhongyi Zhou$^{2}$, Tianyu Xu$^{2}$, Adarsh Kowdle$^{2}$, \\ Inki Kim$^{1}$ , James M. Rehg$^{1}$\thanks{Corresponding author} \\
$^1$ University of Illinois Urbana-Champaign \quad $^2$ Google\\
{\tt\small \{xucao2,jrehg\}@illinois.edu}
}

\begin{document}
\maketitle
\begin{abstract}
Estimating human gaze targets from images in-the-wild is an important and formidable task. Existing approaches primarily employ brittle, multi-stage pipelines that require explicit inputs, like head bounding boxes and human pose, in order to identify the subject of gaze analysis. As a result, detection errors can cascade and lead to failure. Moreover, these prior works lack the flexibility of specifying the gaze analysis task via natural language prompting, an approach which has been shown to have significant benefits in convenience and scalability for other image analysis tasks. 
To overcome these limitations, we introduce the \textbf{Promptable Gaze Target Estimation (PGE)} task, a new end-to-end, concept-driven paradigm for gaze analysis. PGE conditions gaze prediction on flexible user text or visual prompts (e.g., "the boy in the red shirt" or "person in point [0.52, 0.48]") to identify a specific subject for gaze analysis. This approach integrates subject localization with gaze estimation, and eliminates the rigid dependency on intermediate analysis stages. We develop a scalable data engine to generate \textbf{Gaze-Co} (Gaze Estimation with Concepts), a dataset and benchmark of 120K high-quality, prompt-annotated image pairs.
We also propose \textbf{GazeAnywhere}, the first model designed for PGE. GazeAnywhere uses a transformer-based detector to fuse features from frozen encoders and simultaneously solves subject localization, in/out-of-frame presence, and gaze target heatmap estimation. GazeAnywhere achieves state-of-the-art performance on multiple PGE benchmarks, setting a strong baseline for this new problem even on a difficult out-of-domain, real-world clinical dataset. GazeAnywhere is open-sourced in \href{https://github.com/IrohXu/GazeAnywhere}{github.com/IrohXu/GazeAnywhere}.
\end{abstract}


\section{Introduction}

Human gaze is a fundamental non-verbal cue, conveying a wealth of social and cognitive information~\cite{henderson2003human,okumura2013power}. It is integral to human interaction, used to initiate social contact, signal attention and interest, manage conversational turn-taking, and regulate intimacy. As a direct proxy for cognition, gaze can also reveal a person's intentions, preferences, and emotional states~\cite{recasens2015they,tafasca2024toward}. Consequently, its study has attracted significant interest across diverse fields, including psychology, human-computer interaction, and clinical research on conditions such as autism spectrum disorder (ASD)~\cite{argyle1994gaze,falck2012gaze,admoni2017social}.

\begin{figure}[t]
    \centering
    \includegraphics[width=1.0\linewidth]{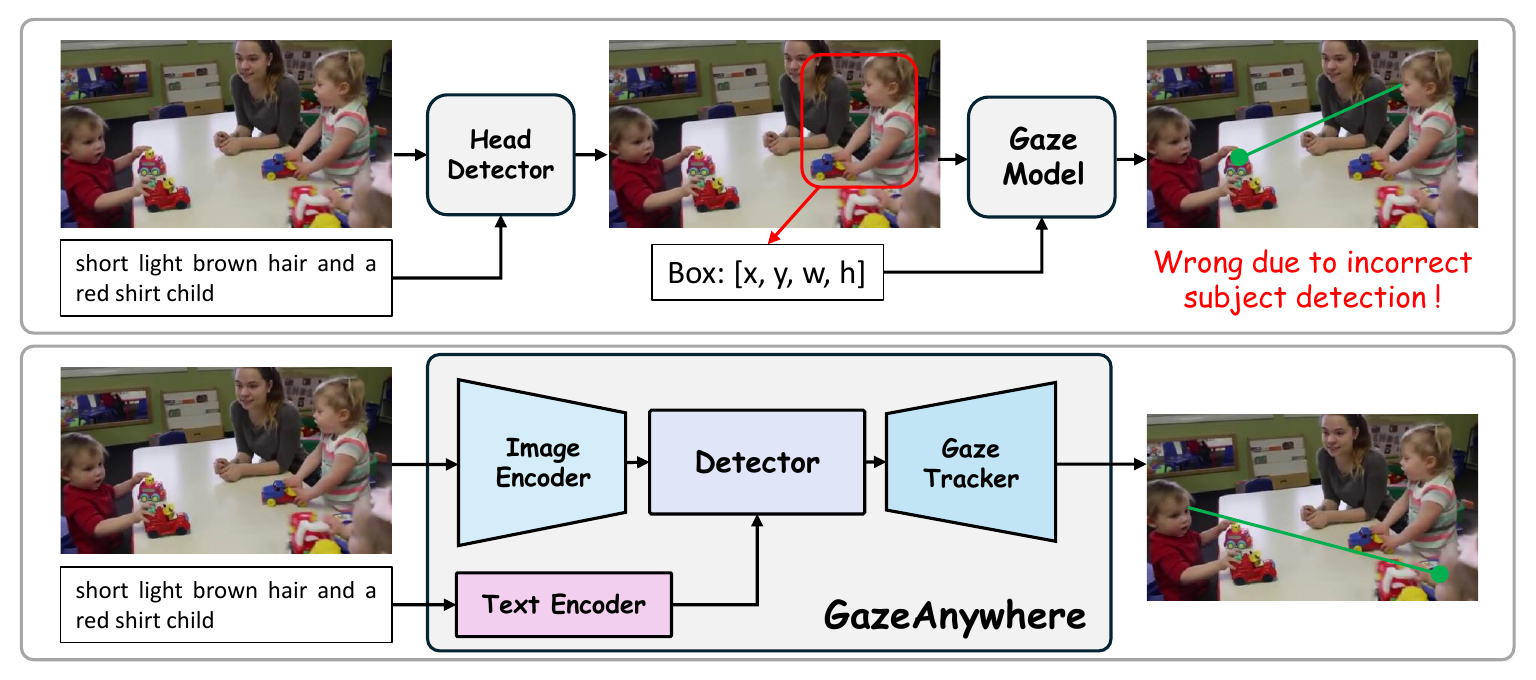}
    \caption{Gaze target estimation in in-the-wild environments. Prior methods such as Sharingan and ViTGaze have to rely on an Open-Vocabulary Detector (OVD) to produce auxiliary head boxes via dynamic human prompts, introducing a sequential dependency that becomes a major bottleneck.}
    \label{fig:intro}
    \vspace{-4mm}
\end{figure}

Despite its importance, accurately estimating a person's gaze target in unconstrained, ``in-the-wild'' settings via image analysis remains a formidable challenge. Current approaches typically require explicit prior information (such as human head and face bounding boxes, pose estimations, and depth) not only for training but also as critical inputs during inference~\cite{ryan2025gaze,song2024vitgaze}. Consequently, the most common architecture is a multi-stage pipeline where intermediate inputs are generated by a series of pre-processing steps prior to gaze estimation (Figure~\ref{fig:intro}). This process typically requires first detecting and tracking a person and then precisely localizing their head or face with a bounding box before the gaze direction can be computed~\cite{cheng2024appearance,zhang2025tcnet}. This sequential dependency creates a critical bottleneck, as inaccuracies in these initial stages (common in crowded scenes, poor lighting, or with challenging cases like detecting children's faces) can cascade, leading to a failure of the entire system.

To overcome this limitation and provide greater convenience and flexibility in specifying gaze analysis tasks, we propose a paradigm shift toward an end-to-end, concept-based framework for gaze target estimation. This approach is inspired by recent advancements in vision foundation models, such as the Open-vocabulary Detectors (OVDs)~\cite{fu2025llmdet,zhang2025sec}, and the Segment Anything Model (SAM) series ~\cite{carion2025sam}, which demonstrate remarkable abilities to detect and segment objects based on semantic-level text or visual concept prompts rather than explicit localization cues via bounding boxes. We extend this core idea to the domain of human gaze understanding, designing a promptable model that can directly identify the gaze target of a specified person within an image. By conditioning on a concept of the subject (e.g., ``the boy in the red shirt''), our approach eliminates the dependencies on intermediate head bounding boxes or pose keypoints. This allows the model to infer gaze targets from a semantic understanding of the scene in a user-friendly end-to-end fashion, paving the way for more robust and versatile systems. Our main contributions can be summarized as:

\begin{itemize}[itemsep=0.5ex, parsep=0pt, topsep=0pt, leftmargin=0.7cm]
\item[\textbf{(1)}] We define the \textbf{Promptable Gaze Target Estimation (PGE)} task, which extends the traditional gaze target estimation problem to an unconstrained, text promptable end-to-end paradigm.

\item[\textbf{(2)}] We design a scalable data engine to generate 120K high quality PGE training annotations consisting of subject text to gaze alignment data pairs.

\item[\textbf{(3)}] We introduce \textbf{GazeAnywhere}, the first promptable concept-driven gaze target estimation model. Our model achieves state-of-the-art performance in several benchmarks including an out-of-domain private dataset for autism children's gaze target estimation.

\end{itemize}

\section{Related Work}
\label{sec:related}

\noindent \textbf{Human Gaze Target Estimation.}
Interpreting gaze is a crucial component of human behavior understanding~\cite{zhang2020human,emery2000eyes}. This has motivated the "gaze-following" task, introduced by datasets like GazeFollow~\cite{recasens2015they,recasens2017following}, VideoAttentionTarget~\cite{chong2020detecting}, GOO~\cite{tomas2021goo} and ChildPlay~\cite{tafasca2023childplay}, which requires a model to predict the scene location a person is looking at. Dominant approaches have thus far employed multi-branch, fusion-based architectures. These models separately process explicit cues such as head position~\cite{yang2024gaze2,song2024vitgaze,yang2024gaze,ryan2025gaze}, pose~\cite{bao2022escnet}, text-based directions~\cite{wang2023gazeclip,miao2024diffusion}, facial expressions~\cite{lai2025clipgaze}, and depth~\cite{tonini2022multimodal,jin2022depth,horanyi2023they,miao2023patch}, subsequently combining these features to predict gaze points~\cite{zhang2015appearance,jin2021multi,cheng2022gaze,zhang2022gazeonce,hu2022gaze,miao2023patch,tu2022end,lin2025gazehta}. In addition, existing end-to-end models cannot specify the subject person during inference either~\cite{tonini2023object,tu2022end,de2025gazedetr}. While effective, these strategies are dependent on the availability and accuracy of these intermediate representations. 
Concurrently, other studies have expanded the task's scope, such as multi-view gaze target estimation~\cite{miao2025multi} and GazeHOI~\cite{tafasca2024toward} for open-vocabulary targets. Despite these advances, a common limitation still persists: a dependency on auxiliary information, such as precise head bounding boxes or pose estimations. This reliance poses significant challenges in realistic settings, where such priors are often unreliable or unavailable.

\noindent \textbf{Promptable and Interactive Object Detection.} Recent advances in promptable and open-vocabulary perception are enabling models to generalize beyond fixed label sets, especially the referring expression comprehension tasks~\cite{carion2020end,yao2022detclip,zhang2023text,minderer2023scaling,niu2024overview,jiang2025referring}. OVD methods, for instance, leverage large-scale vision-language encoders like CLIP to detect arbitrary concepts specified by text at inference time, even for categories unseen during training~\cite{zhou2022detecting,liu2024grounding,minderer2022simple,fu2025llmdet}. In parallel, interactive segmentation frameworks demonstrate how models can respond to flexible text and visual prompts~\cite{lai2024lisa,carion2025sam}. Together, these advances suggest the new paradigm: visual perception systems can be effectively guided by conceptual cues, rather than rigidly predefined supervision. Building on this paradigm, we formulate gaze target estimation as a concept-conditioned reasoning task, where both the subject and their attended region are inferred from semantic prompts rather than explicit localization inputs.

\noindent \textbf{Vision Foundation Models.} 
Vision Foundation Models (VFMs) have become a dominant approach in computer vision, entailing a significant shift of models trained on massive web-scale datasets. VFMs primarily include two branches: (1) weakly-supervised models like CLIP~\cite{radford2021learning}, SigLIP~\cite{zhai2023sigmoid,tschannen2025siglip}, and MetaCLIP~\cite{xu2023demystifying,chuang2025meta,bolya2025perception}, which learn powerful representations from image-text pairs using contrastive losses, and (2) self-supervised learning (SSL) models like DINO series~\cite{jose2025dinov2,simeoni2025dinov3,jose2025dinov2}, which learn robust visual features from unlabeled images. These powerful, pre-trained encoders now serve as general-purpose backbones for a wide range of downstream tasks. This trend has also influenced gaze estimation, where systems like ViTGaze~\cite{song2024vitgaze} and Gaze-LLE~\cite{ryan2025gaze} have successfully adapted VFM architectures and leveraged their pre-trained features to improve performance, demonstrating the value of these models for fine-grained, human-centric tasks.

\section{Method}

\begin{figure*}[t]
    \centering
    \includegraphics[width=0.95\linewidth]{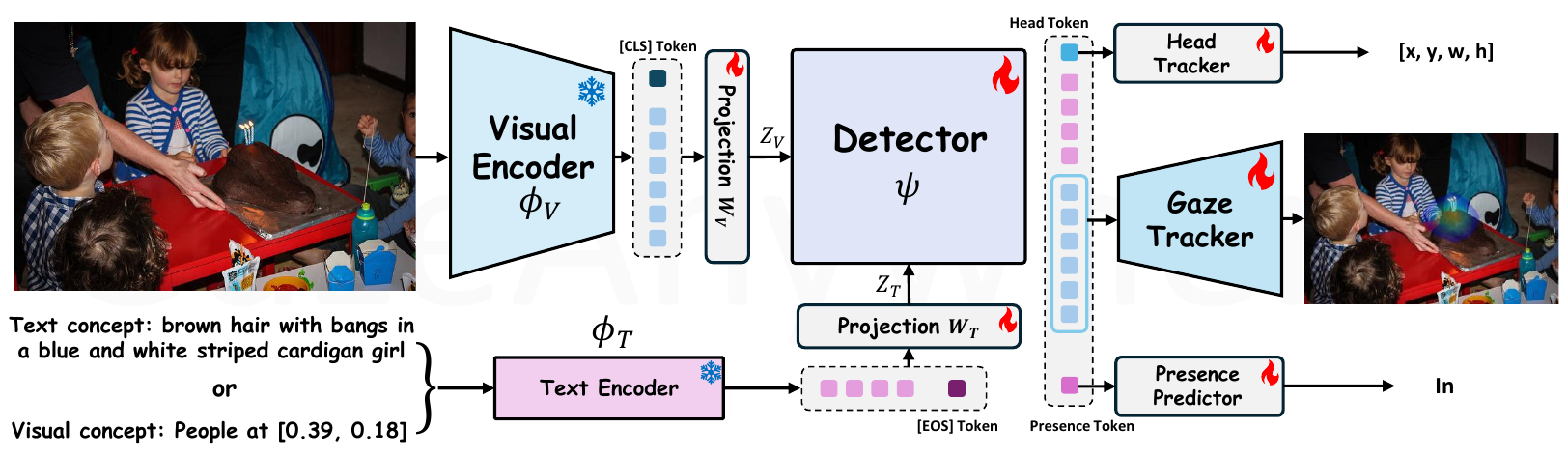}
    \caption{An overview of the GazeAnywhere end-to-end framework for PGE. The model uses frozen visual (DINOv3) and text (dino.txt) encoders to provide features to a trainable transformer-based detector. This detector utilizes multiple decoders to simultaneously predict a subject's head bounding box , in-frame presence , and gaze target. GazeAnywhere is conditioned on flexible user prompts, such as a natural language text description or visual (coordinate-based) cues.}
    \label{fig:framework}
    \vspace{-2mm}
\end{figure*}

\subsection{Promptable Gaze Target Estimation}

We define the PGE task as estimating the gaze target location of a specific subject within an image or video, identified by a user-provided prompt. This prompt can be of two types: text prompting via a short, natural language query; and visual one by prompting a spatial coordinate, such as the center point of a head bounding box. In text prompting, our goal is to support any simple, visually-groundable noun phrase as a text prompt. However, this introduces intrinsic ambiguity (e.g., "the person in the back"). To mitigate this and enable unambiguous identification, we structure text prompts around four main categories. A user can combine descriptions from these categories to specify a subject: (1) Appearance: Noun phrases describing a person, consisting of an identity (e.g., woman, man, child) and optional modifiers (e.g., hair type/color, clothes, glasses, hat). (2) Location: The subject's spatial position in the image (e.g., center, left, top-right). (3) Pose: The subject's static posture. (4) Action: Verb phrases describing what the subject is doing.

More formally, given an input RGB image $I \in \mathbb{R}^{3 \times H \times W}$ and a prompt $P$ (either text $T$ or a visual cue), the goal of PGE is to produce a gaze heatmap $\hat{H} \in \mathbb{R}^{H_{\text{out}} \times W_{\text{out}}}$. Each element $\hat{H}(i, j)$ represents the probability that the subject specified by $P$ is gazing at the spatial location $(i, j)$. Unlike classic gaze estimation methods, PGE demands that the model solve the task end-to-end, directly linking a flexible, high-level query to a final gaze heatmap. This formulation is substantially more challenging as it precludes the use of auxiliary inputs common in traditional pipelines (e.g., subject bounding boxes, pose keypoints, or depth maps). Our model must implicitly learn to perform subject identification, localization, and gaze estimation jointly with text input, rather than relying on the explicit outputs of separate, specialized models like open-vocabulary detectors or pose estimators.

\subsection{GazeAnywhere Architecture}

Figure~\ref{fig:framework} shows the overall architecture of GazeAnywhere. The model consists of a frozen image encoder, a frozen text encoder to proceed visual modality and text modality. A transformer-based detector is used to learn joint representations and map text prompt into the main gaze target estimation and auxiliary tasks. 

\vspace{-4mm}
\paragraph{Image Encoder.}
We use a frozen ViT, denoted $\phi_V(\cdot)$, as our image encoder to extract general visual features. Consistent with PGE's problem definition, we do not employ any auxiliary models for dedicated depth or pose feature extraction. The image encoder processes an input image $I \in \mathbb{R}^{3 \times H \times W}$ by dividing it into a sequence of $N_V$ patch tokens, to which a learnable \texttt{[CLS]} token is prepended. The resulting output sequence from $\phi_V$ is $\phi_V(I)=[ c, s_1, s_2, \dots, s_{N_V} ] \in \mathbb{R}^{(1+N_V) \times D_{V}}$. Here, $D_{V}$ is the embedding dimension, $c \in \mathbb{R}^{D_{V}}$ is the final \texttt{[CLS]} token embedding, and $s_i \in \mathbb{R}^{D_{V}}$ is the output embedding for the $i$-th patch. 

\vspace{-4mm}
\paragraph{Text Encoder.}
We employ a frozen text encoder, $\phi_T(\cdot)$, which consists of a series of transformer blocks and a final linear layer. The linear layer maps the output \texttt{[EOS]} token's feature to the image embedding space.
To prepare the input, a tokenizer first converts the text $T$ into a sequence of token IDs. These IDs are then mapped to initial text embeddings $T_E$ via an embedding layer, and the sequence is padded to a fixed context length, $L_T$.
The encoder $\phi_T$ processes this embedding sequence $T_E$, producing the final output sequence: $\phi_T(T_E)=[t_1, t_2, \dots, t_{N_T}, t_{eos}, \dots, t_{pad}] \in \mathbb{R}^{L_T \times D_{T}}$. Here, $D_{T}$ is the output embedding dimension, $t_i \in \mathbb{R}^{D_{T}}$ is the output embedding for the $i$-th content token, and $t_{eos}, t_{pad} \in \mathbb{R}^{D_{T}}$ are the special tokens representing the end of the input and padding, respectively.

\vspace{-4mm}
\paragraph{Projection Layers.}
The image encoder $\phi_V$ and text encoder $\phi_T$ output features with potentially different dimensions, $D_V$ and $D_T$, respectively. To map these features into a unified, shared space, we introduce two trainable linear projection layers, $W_V$ and $W_T$. These layers project the high-dimensional features into a common, lower-dimension $D$, where $D < \min(D_V, D_T)$:

\begin{equation}
    Z_{\text{V}} = W_{V} \cdot \phi_V(I) \label{eq:proj_v}
\end{equation}
\begin{equation}
    Z_{\text{T}} = W_{T} \cdot \phi_T(T_E) \label{eq:proj_t}
\end{equation}

Here, $W_V$ and $W_T$ are the learnable projection matrices, and the operation \texttt{$\cdot$} denotes a per-token linear transformation. This process results in a sequence of visual tokens $Z_{\text{V}} \in \mathbb{R}^{(1+N_V) \times D}$ and text tokens $Z_{\text{T}} \in \mathbb{R}^{L_T \times D}$, which now share the same embedding dimension.

\paragraph{Task-Specific Embeddings.}
Beyond the primary cross-modal feature alignment, we introduce two specialized, learnable embeddings to explicitly model key sub-problems: a \textbf{head token} and a \textbf{gaze presence token}.

\begin{enumerate}
    \item \textbf{Head Token:} This is a learnable embedding designed to explicitly predict the head localization of the prompted subject, serving as the image-text alignment objective in our task. It is initialized using the embedding of the text \texttt{[EOS]} token.
    
    \item \textbf{Target Presence Token:} This token is introduced to address the \textit{in/out-of-frame} gaze target boolean prediction objective. The rationale for this is that the in/out decision relies on global contextual cues from the entire image, which conflicts with the inherently local nature of the target localization objective. Forcing a single query or mechanism to handle both can be counterproductive. Therefore, we decouple the localization and in/out prediction tasks. This dedicated, learnable global token is responsible for the in/out prediction and is initialized using the embedding of the visual \texttt{[CLS]} token.
\end{enumerate}

\vspace{-3mm}
\paragraph{Detector Transformer.}
After extracting and projecting the visual and text features, we introduce a \textbf{Detector Transformer}, $\psi(\cdot)$, to fuse these representations and refine them for the gaze target estimation task. $\psi(\cdot)$ apply the same Transfomrer block in DINOv3~\cite{simeoni2025dinov3}. The input to $\psi$ is a single sequence $F$ constructed by concatenating the projected features and our specialized task tokens.

First, we define the \textbf{head token} $t_h \in \mathbb{R}^{D}$ and \textbf{target presence token} $t_p \in \mathbb{R}^{D}$. These are formed by combining the projected global tokens ($c'$ from vision, $t'_{eos}$ from text) with dedicated learnable embeddings, $\text{E}_{\text{presence}}$:

\begin{equation} 
    t_h = t'_{eos}  \label{eq:head_token}
\end{equation} 
\begin{equation} 
    t_p = c' + \text{E}_{\text{presence}} \label{eq:presence_token} 
\end{equation}

Let $\mathbf{s}' = [s'_{1}, \dots, s'_{N_V}]$ be the sequence of projected visual patch tokens from $Z_V$ (excluding $c'$) and $\mathbf{t}' = [t'_1, \dots, t'_{N_T}]$ be the projected text content tokens from $Z_T$ (excluding $t'_{eos}$ and padding). The full input sequence $F$ is then assembled as:

\begin{equation}
    F = [t_h, \mathbf{t}', \mathbf{s}', t_p] \in \mathbb{R}^{(N_T + N_V + 2) \times D} \label{eq:full_seq}
\end{equation}

We inject positional information by adding 1D sinusoidal position embeddings to the text tokens $\mathbf{t}'$ and 2D sinusoidal position embeddings to the visual tokens $\mathbf{s}'$~\cite{dosovitskiy2020image}. The Detector transformer $\psi$ is a stack of $k$ standard transformer blocks; $k$ is a hyperparameter ablated in our experiments. $\psi$ processes $F$ and outputs a refined sequence of the same dimension, $\psi(F) \in \mathbb{R}^{(N_T + N_V + 2) \times D}$. Specific tokens from this output are then passed to dedicated decoders.

\vspace{-3mm}
\paragraph{Decoders.}
The Detector transformer $\psi$ outputs a refined sequence of tokens. We attach three distinct prediction heads to specific tokens from this sequence to produce the final outputs.

\begin{itemize}
    \item \textbf{Gaze Tracker (Heatmap Decoder):} The refined visual patch tokens $\mathbf{\hat{s}} \in \mathbb{R}^{N_V \times D}$ are first re-assembled from their 1D sequence form back into a 2D spatial grid. This feature map is then fed through a convolutional decoder, consisting of two transposed convolutional layers, which upsamples the features to the output heatmap $\hat{H} \in \mathbb{R}^{H_{out} \times W_{out}}$. In our experiments, we set $H_{out} = W_{out} = 64$.

    \item \textbf{Head Tracker (Box Decoder):} We use the refined head token $\hat{t}_h \in \mathbb{R}^{D}$ for an auxiliary head localization task. The token is passed through a 3-layer feed-forward network (FFN) with ReLU activations and a hidden dimension of $D$. This head regresses a 4-dimensional vector $[x, y, w, h]$ representing the normalized center coordinates, width, and height of the subject's head box.

    \item \textbf{Presence Predictor (In/Out Decoder):} The refined gaze presence token $\hat{t}_p \in \mathbb{R}^{D}$ is used to predict whether the gaze target is in or out of the frame. It is processed by a 2-layer FFN (with one hidden layer of dimension $D$ and ReLU activation) that outputs a single logit for the binary classification.
\end{itemize}

\subsection{Learning Objective}
We train our model end-to-end with a joint multi-task objective. The total loss $\mathcal{L}_{\text{total}}$ is a weighted linear combination of three loss terms: one for the gaze heatmap, one for the gaze presence, and one for the auxiliary head localization task.

\begin{equation}
\mathcal{L}_{\text{total}} =  \mathcal{L}_{\text{gaze}} + \mathcal{L}_{\text{presence}} + \mathcal{L}_{\text{head}}
\label{eq:total_loss}
\end{equation}

The gaze heatmap loss $\mathcal{L}_{\text{gaze}}$ is a pixel-wise binary cross-entropy (BCE) loss. The supervisory target is a heatmap $Y$, constructed by placing a 2D Gaussian ($\sigma=3$) at the ground-truth gaze target location. Let $\hat{Y}$ be the predicted heatmap. The loss is defined as:

\begin{equation}
\mathcal{L}_{\text{gaze}} = - \frac{1}{N} \sum_{p=1}^{N} \left[ y_p \log(\hat{y}_p) + (1 - y_p) \log(1 - \hat{y}_p) \right]
\label{eq:gaze_loss}
\end{equation}

where $N = H_{out} \times W_{out}$ is the total number of pixels, and $y_p$ and $\hat{y}_p$ are the ground-truth and predicted values for a single pixel $p$, respectively.

The gaze presence loss $\mathcal{L}_{\text{presence}}$ is a Focal Loss supervised with a binary label $Y_{\text{presence}} \in \{0, 1\}$. Let $\hat{Y}_{\text{presence}} \in [0, 1]$ be the model's predicted probability that the target is present ($Y_{\text{presence}} = 1$). The loss is defined as:

\begin{equation}
\mathcal{L}_{\text{presence}} = \mathcal{L}_{\text{focal}}(Y_{\text{presence}},\hat{Y}_{\text{presence}})
\label{eq:presence_loss}
\end{equation}

where the hyperparameter of the focal loss is the default value from ~\cite{lin2017focal}.






The subject head bounding box loss $\mathcal{L}_{\text{head}}$ is a linear combination of the $\mathcal{L}_1$ loss and the generalized IoU loss. which is widely used by object detection tasks. It defined as:

\begin{equation}
\mathcal{L}_{\text{head}} = \lambda_{l_1} || b-\hat{b} ||_1 + \lambda_{\text{iou}} \mathcal{L}_{\text{iou}} (b, \hat{b})
\end{equation}

where $b$ and $\hat{b}$ is the ground truth head box and predicted head box. $\mathcal{L}_{\text{IoU}}$ is the GIoU loss~\cite{rezatofighi2019generalized,carion2020end}. $\lambda_{l_1}$ and $\lambda_{\text{iou}}$ are head object detection hyperparameters. We followed DETR~\cite{carion2020end} and OWLViT~\cite{minderer2022simple} to set $\lambda_{l_1}=5$ and $\lambda_{\text{iou}}=2$.

\section{Gaze with Concept (Gaze-Co) Dataset}

Training GazeAnywhere for the PGE task requires a large and diverse dataset annotated with concepts, a resource that no existing gaze dataset provides. To address this, we developed a scalable data engine that generates annotations via a human-in-the-loop feedback process. This engine worked in tandem with two human annotators (co-authors) to perform several key functions: aligning heterogeneous annotations, filtering low-quality frames, generating concise concept phrases, and facilitating human verification. After three rounds of iteration, we created Gaze-Co, the first large-scale dataset for PGE, containing 120K samples sourced from the training set of GazeFollow, VisualAttentionTarget (VAT) and ChildPlay. To establish a comprehensive benchmark, we also converted the test sets of these well-known gaze datasets to the PGE format, creating GazeFollow-Concept, VAT-Concept, and ChildPlay-Concept. We further conducted experiments on a private, Institutional Review Board (IRB)-approved, out-of-domain (OOD) evaluation set with several frames in 40 child social communication (Child-SC) videos.

\subsection{Data Engine}

\begin{figure}[t]
    \centering
    \includegraphics[width=1.0\linewidth]{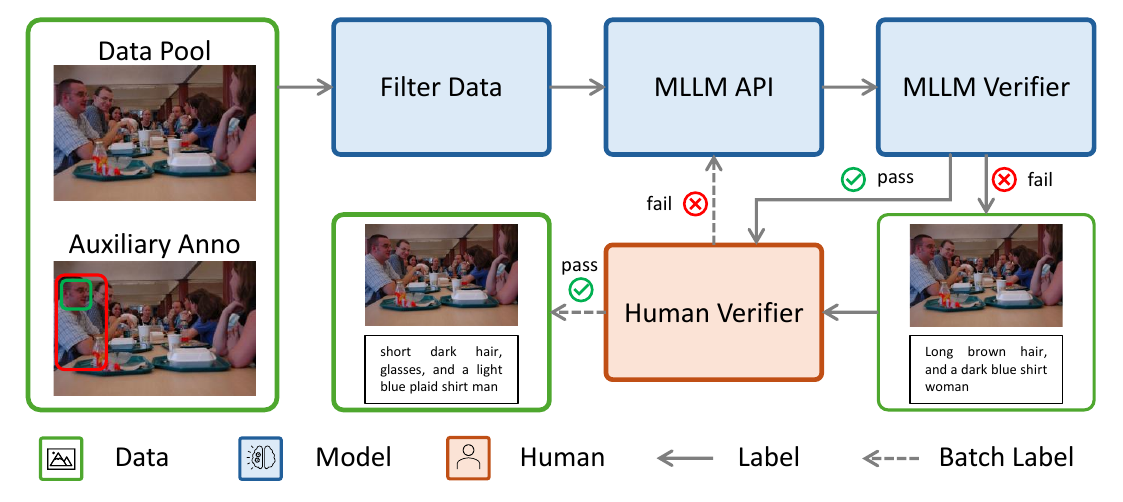}
    \caption{Overview of the GazeAnywhere data engine.}
    \label{fig:data_engine}
    \vspace{-3mm}
\end{figure}

Figure~\ref{fig:data_engine} illustrates the workflow of the data engine. We can divide the process into three stages: (1) data alignment and filter; (2) concept generation; (3) verification.

\noindent \textbf{Data Alignment and Filter.} 
The source datasets differ in coordinate conventions, split policies, and metadata. We therefore adopt a unified schema with explicit pixel coordinates for the head box $(x_{min}, y_{min}, x_{max}, y_{max})$, and a normalized gaze point $(g_x/W, g_y/H)$. Then, to ensure reliable person-scale learning signals, we apply geometric and sharpness filters. Annotations are retained only if the head box width~$\geq$~30\,px, height~$\geq$~40\,px, area~$\geq$~2500\,px$^2$, and the box-to-image ratio~$\in[0.008,\,0.3]$, with sufficient Tenengrad focus. These thresholds remove extremely small, oversized, or blurry instances while preserving diverse valid samples.

\noindent \textbf{Concept Generation.} 
For each retained annotation, we produce a short, lowercase concept phrase comprising attribute, position, action, and pose, together with a coarse count of visible people. Concept generation is executed with a production Vision Language Model (VLM) accessed through API (Gemini 2.5 Pro \cite{google_gemini_api_2025,comanici2025gemini}), using batch processing with fixed prompts and rates. The attribute phase privileges stable visual cues (hair, glasses, beard, colors, and patterns) and the final token is constrained to one of {man, woman, boy, girl, infant, child} as an apparent (perceived) age/sex presentation label, used solely as a visual category cue rather than a verified identity attribute; when indeterminate, we write “adult” or “child. The position uses brief canvas references (e.g., “bottom left corner”). Action and pose are explicitly non-overlapping: action describes ongoing interaction or motion with object or direction when visible, while pose captures static body configuration and facing direction. When a field is indeterminate, we write “none.” 

\noindent \textbf{Verification.}
We adopt an Multi-modal Large Language Model (MLLM)-first, human-in-the-loop verification workflow. The Gemini 2.5 Pro reviews all generated concepts and flags each as pass or fail. Human annotators then spot-check a random subset of the MLLM passed cases and evaluate the batch success rate. During review, both the MLLM and human annotators check whether each concept correctly matches its designated head box (consistency), whether all four fields are present and non-conflicting (completeness), and whether the text contains no sensitive or identifying information (privacy). If the human verifier finds the batch success rate is low, the data engine will return to the concept generation stage. The human verifier then adjusts the prompts and rules and re-runs the concept generation and verification until the observed error rate is kept low ($\leq 1\%$). For the private Child-SC dataset with IRB restriction, all concept annotations are generated manually by authorized human annotators without being sent to the MLLM.

\begin{table*}[t]
  \centering

{\fontsize{8pt}{11pt}\selectfont
   \resizebox{1.00\linewidth}{!}{%
     \begin{NiceTabular}{l l c c c c c c c c c c c}[colortbl-like]
       \toprule[1.2pt]
       \multirow{2}{*}{\textbf{Gaze Model}} & \multirow{2}{*}{\textbf{Head Detection}} & \multirow{2}{*}{\textbf{\#Total Param}} & \multirow{2}{*}{\textbf{Time per sample[ms]}\(\downarrow\)} & \multicolumn{3}{c}{\textbf{GazeFollow-Concept}} & \multicolumn{2}{c}{\textbf{VAT-Concept}} & \multicolumn{2}{c}{\textbf{ChildPlay-Concept}} & \multicolumn{2}{c}{\textbf{Child-SC (OOD)}} \\ \cmidrule(lr){5-7} \cmidrule(lr){8-9} \cmidrule(lr){10-11} \cmidrule(lr){12-13}
       & & & & AUC \(\uparrow\) & Avg L2 \(\downarrow\) & Min L2 \(\downarrow\) & L2 \(\downarrow\) & AP \(\uparrow\) & L2 \(\downarrow\) & AP \(\uparrow\) & L2 \(\downarrow\) & AP \(\uparrow\) \\
       \midrule[1.2pt]
       \multirow{5}{*}{ViTGaze~\cite{song2024vitgaze}} & Ground-Truth & - & - & 0.955 & 0.097 & 0.045 & 0.098 & 0.879 & 0.113 & 0.905 & - & - \\
        & GroundingDINO-B~\cite{liu2024grounding} & 255M & 116 & 0.913 &  0.192 & 0.116 & 0.232 & 0.792 & 0.170 & 0.863 & 0.139 & 0.795 \\
        & LLMDet-L~\cite{fu2025llmdet} & 365M & 339 & 0.897 & 0.189 & 0.113 & 0.235 & 0.763 & 0.158 & 0.869 & 0.165 & 0.825 \\
        & OWLv2-L~\cite{minderer2023scaling} & 460M & 179 & 0.927 & 0.170 & 0.094 & 0.241 & 0.749 & 0.151 & 0.862 & 0.168 & 0.781 \\
        & RexSeek~\cite{jiang2025referring} & 3B & 1160 & 0.945 & 0.119 & 0.063 & 0.123 & 0.841 & 0.117 & 0.903 & 0.136 & 0.818\\
       \midrule
       \multirow{5}{*}{Sharingan~\cite{tafasca2024sharingan}} & Ground-Truth & - & - & 0.944 & 0.114 & 0.059 & 0.109 & 0.852 & 0.118 & 0.854 & - & -\\
        & GroundingDINO-B~\cite{liu2024grounding} & 338M & 119 & 0.860 & 0.259 & 0.178 & 0.352 & 0.720 & 0.216 & 0.888 & 0.349 & 0.771  \\
        & LLMDet-L~\cite{fu2025llmdet} & 449M & 342 & 0.865 & 0.241 & 0.160 & 0.281 & 0.785 & 0.213 & 0.874 & 0.347 & 0.832 \\
        & OWLv2-L~\cite{minderer2023scaling} & 544M & 182 & 0.867 & 0.255 & 0.173 & 0.291 & 0.745 & 0.185 & 0.893 & 0.263 & 0.847 \\
        & RexSeek~\cite{jiang2025referring} & 3B  & 1159 & 0.888 & 0.207 & 0.142 & 0.273 & 0.601 & 0.178 & 0.863 & 0.166 & 0.810\\
        \midrule
       \multirow{5}{*}{Gaze-LLE~\cite{ryan2025gaze}} & Ground-Truth & - & - & 0.961 & 0.099 & 0.045 & 0.101 & 0.875 & 0.113 & 0.912 & - & - \\
        & GroundingDINO-B~\cite{liu2024grounding} & 539M & 145 & 0.925 & 0.165 & 0.106 & 0.234 & 0.780 & 0.152 & 0.863 & 0.172 &  0.856 \\
        & LLMDet-L~\cite{fu2025llmdet} & 650M & 368 & 0.912 & 0.168 & 0.109 & 0.230 & 0.793 & 0.145 & 0.875 & 0.175 & 0.855\\
        & OWLv2-L~\cite{minderer2023scaling} & 745M & 208 & 0.941 & 0.146 & 0.089 & 0.229 & 0.792 & 0.127 & 0.893 & 0.161 & 0.830 \\
        & RexSeek~\cite{jiang2025referring} & 3B & 1183 & 0.954 & 0.108 & 0.054 & \textbf{0.121} & 0.861 & 0.119 & 0.914 & 0.172 & 0.846 \\
       \midrule
       \multicolumn{2}{c}{\rowcolor{gray!20} \textbf{\modelnamenc-CLIP-L}} & 430M & 35 & 0.953 & 0.105 & 0.056 & 0.137 & 0.874 & 0.104 & \textbf{0.915} & 0.146 & 0.868 \\
       \multicolumn{2}{c}{\rowcolor{gray!20} \textbf{\modelnamenc-DINOv3-L}} & 870M & 96 & \textbf{0.958} & \textbf{0.099} & \textbf{0.050} & 0.123 & \textbf{0.879}  & \textbf{0.098} & 0.906 & \textbf{0.090} &  \textbf{0.902} \\
       \bottomrule
       \end{NiceTabular}%
   }
  }
\caption{PGE results on four datasets. The input is the text prompt of the subject person's appearance, position, action and pose. For baseline methods, the OVD is used to extract the bounding box with the input prompt and then feed the bounding box to the gaze models. Latency is compared with inference running speed per image in \textit{batch size} = 1.}
  \label{tab:main_experiment}
  \vspace{-2mm}
\end{table*}

\subsection{Gaze-Co Dataset and Benchmarks}

Gaze-Co is the first large-scale dataset for promptable gaze target estimation, unifying GazeFollow, VAT, and ChildPlay under a shared schema with concept-level annotations for both training and evaluation.

\noindent \textbf{Training Data.} The Gaze-Co 120K training set contains about 120K images from the official training splits of the three source datasets. Each record includes the target head box, normalized gaze point, in/out-of-frame label, and a compact concept phrase (attribute, position, action, and pose). All samples pass image quality filters, ensuring diverse, valid instances across viewpoint, poses, scales, and interaction contexts.

\noindent \textbf{Benchmark Settings.} The benchmark uses the official test splits of GazeFollow, VAT, and ChildPlay, each converted into the Gaze-Co format. We evaluate concept-conditioned gaze prediction under three settings: (i) in-domain testing on the test sets; and (ii) OOD evaluation on the  Child-SC dataset, a  developmental sample of children's gaze behavior collected under an IRB-approved study (see the appendix for the dataset description). Every text prompt in the test set has been human-verified rather than spot-checked, ensuring accuracy and consistency. Each model receives the image, with the concept text added or altered according to the test setting. This setup provides a consistent framework for comparing models under controlled concepts-based conditions.

\subsection{Metrics}
We evaluate models using heatmap Area Under the Curve (AUC) in GazeFollow and pixelwise L2 in all. For heatmap AUC, the predicted heatmap is treated as a confidence map to compute an ROC curve against the binary gaze target map. Pixelwise L2 measures the Euclidean distance between the heatmap peak and the ground-truth gaze point. For GazeFollow-Concept, each image includes multiple gaze annotations directed at the same target person, so we additionally report Avg L2 (distance to the mean of all targets) and Min L2 (distance to the nearest target). For VAT-Concept, ChildPlay-Concept, and Child-SC (the IRB-approved OOD set), annotations include binary in/out labels relative to the target region; thus, we report pixelwise L2, and average precision (AP) to jointly evaluate localization and in/out binary classification.

\section{Experiments}

We evaluate GazeAnywhere on the PGE task, comparing its text-prompting capabilities against State-of-the-Art (SOTA) two-stage pipelines that integrate OVDs for head/human detection with a separate gaze modeling stage. We also present a series of ablation studies demonstrating the importance of the frozen encoders, validating our loss design, and analyzing the differences between visual and text prompting. Finally, we demonstrate a real-world application, the ``AnyGaze Agent,'' a system that integrates GazeAnywhere with an Augmented Reality (AR) device and a MLLM.

\subsection{Implementation Details}

All models are trained for 25 epochs using the Adam optimizer and a cosine learning rate scheduler with an initial rate of 1e-3 and batch size 128, followed by an additional 5 epochs with a reduced learning rate of 1e-5. All training experiments are conducted with 4 NVIDIA H100 GPUs. The inference is running with 1 NVIDIA L40S GPU. We adopt the DigiLens ARGO smartglass as the AR platform to deploy AnyGaze Agent for real world experiments. More details are shown in the appendix.

\subsection{Main Results}

Table~\ref{tab:main_experiment} compares GazeAnywhere against strong two-stage baselines, which we created by pairing three SOTA gaze methods Gaze-LLE~\cite{ryan2025gaze}, Sharingan~\cite{tafasca2024sharingan}, ViTGaze~\cite{song2024vitgaze} with three leading OVDs for human detection~\cite{liu2024grounding,fu2025llmdet,minderer2023scaling}. Details of these baselines are shown in the appendix. The encoders of GazeAnywhere can be CLIP-L~\cite{radford2021learning} or DINOv3-L~\cite{simeoni2025dinov3} with dino.txt~\cite{jose2025dinov2}. On the PGE text-prompting task, GazeAnywhere achieves SOTA performance on all metrics across the three public datasets, as well as on our challenging OOD private dataset from a real-world assessment setting in which children's social communication skills are quantified by experts.

\subsection{Ablation Study}

We use GazeAnywhere-DINOv3-L for all following up ablation experiments in GazeFollow-Concept and VAT-Concept. More results are shown in Appendix Sec~\ref{appendix:more_results}.



\begin{table}[t]
  \centering
  {\fontsize{8pt}{11pt}\selectfont
   \resizebox{0.99\linewidth}{!}{%
     \begin{NiceTabular}{l c c c c c}[colortbl-like]
       \toprule[1.2pt]
       \multirow{2}{*}{\textbf{Model}} & \multirow{2}{*}{\textbf{Param}} & \multicolumn{2}{c}{\textbf{GazeFollow-Concept}} & \multicolumn{2}{c}{\textbf{VAT-Concept}} \\ \cmidrule(lr){3-4} \cmidrule(lr){5-6}
       & & Avg L2 \(\downarrow\) & Min L2 \(\downarrow\) & L2 \(\downarrow\) & AP \(\uparrow\) \\
       \midrule[1.2pt]
       Qwen3-VL-8B~\cite{Qwen2.5-VL} & 8B &  0.201 & 0.137 & 0.286 & 0.651 \\
       Gemini 2.5 Flash~\cite{comanici2025gemini} & - & 0.216 & 0.156 & 0.292 & 0.661 \\
       \rowcolor{gray!20} GazeAnywhere-DINOv3-L & 870M  & \textbf{0.099} & \textbf{0.050}  & \textbf{0.123} & \textbf{0.879} \\
       \bottomrule
     \end{NiceTabular}%
   }
  }
\caption{Compare GazeAnywhere with SOTA VLMs.}
  \label{tab:ablation_vlm}
\end{table}

\noindent \textbf{Compare GazeAnywhere with SOTA VLMs.} To demonstrate the utility of GazeAnywhere in PGE, we evaluate the 0-shot performance of SOTA VLM on gaze point prediction. Table~\ref{tab:ablation_vlm} shows the comparison of GazeAnywhere, Qwen3-VL-8B and Gemini 2.5 Flash. GazeAnywhere surpass all of them,  highlighting the importance of building specific model for PGE.

\noindent \textbf{PGE with Different Prompting.} GazeAnywhere supports both visual (coordinate-based text) and text (natural language) prompts, as illustrated in Figure~\ref{fig:framework}. In Table~\ref{tab:ablation_prompt}, we compare the performance of these different strategies. We find that text-based prompting achieves performance on par with visual prompting. Furthermore, our decomposition analysis of text prompt composition reveals that the subject's appearance and pose description are the most critical components for the PGE task.

\begin{table}[t]
  \centering
  {\fontsize{8pt}{11pt}\selectfont
   \resizebox{1.00\linewidth}{!}{%
     \begin{NiceTabular}{l c c c c c c c}[colortbl-like]
       \toprule[1.2pt]
       \multirow{2}{*}{\textbf{Prompt type}} & \multicolumn{3}{c}{\textbf{GazeFollow-Concept}} & \multicolumn{3}{c}{\textbf{VAT-Concept}} \\ \cmidrule(lr){2-4} \cmidrule(lr){5-7}
       & AUC \(\uparrow\) & Avg L2 \(\downarrow\) & Min L2 \(\downarrow\) & AUC \(\uparrow\) & L2 \(\downarrow\) & AP \(\uparrow\) \\
       \midrule[1.2pt]
       No prompting & 0.944 & 0.144 & 0.090 & 0.875 & 0.210 & 0.796  \\
       visual prompting & 0.958 & 0.100 & 0.050 & 0.914 & 0.131 & 0.894 \\
       text prompting (appearance) & 0.952 & 0.113 & 0.062 & 0.904 & 0.153 & 0.840 \\
       text prompting (position) & 0.953 & 0.124 & 0.073 & 0.893 & 0.188 & 0.826  \\
       text prompting (action) & 0.949 & 0.129 & 0.078 & 0.897 & 0.180 &  0.839  \\
       text prompting (pose) & 0.952 & 0.121 & 0.070 & 0.909 & 0.163 & 0.859 \\
       text prompting (all) & 0.958 & 0.099 & 0.050 & 0.928 & 0.123 & 0.879 \\
       \bottomrule
     \end{NiceTabular}%
   }
  }
\caption{Comparison of different prompt strategies.}
  \label{tab:ablation_prompt}
  \vspace{-2mm}
\end{table}

\noindent \textbf{Loss ablations.} We conducted an ablation study (Table~\ref{tab:ablation_loss}) on our objective function's components: gaze heatmap, presence, and head losses. The essential gaze heatmap loss was always active, while we trained models removing the presence loss, the head loss, and both. Results indicate the presence loss only supports the auxiliary in/out prediction, not help gaze estimation. The head loss, however, improves both the gaze target estimation and the target presence prediction.

\begin{table}[t]
  \centering
  {\fontsize{8pt}{11pt}\selectfont
   \resizebox{1.00\linewidth}{!}{%
     \begin{NiceTabular}{l l c c c c c c c c}[colortbl-like]
       \toprule[1.2pt]
       \multirow{2}{*}{\textbf{$\mathcal{L}_{\text{gaze}}$}} & \multirow{2}{*}{\textbf{$\mathcal{L}_{\text{presence}}$}} & \multirow{2}{*}{\textbf{$\mathcal{L}_{\text{head}}$}} & \multicolumn{3}{c}{\textbf{GazeFollow-Concept}} & \multicolumn{3}{c}{\textbf{VAT-Concept}} \\ \cmidrule(lr){4-6} \cmidrule(lr){7-9}
       & & & AUC \(\uparrow\) & Avg L2 \(\downarrow\) & Min L2 \(\downarrow\) & AUC \(\uparrow\) & L2 \(\downarrow\) & AP \(\uparrow\) \\
       \midrule[1.2pt]
       \ccmark & \cxmark & \cxmark & 0.956 & 0.102 & 0.052 & 0.925 & 0.135 & - \\
       \ccmark & \ccmark & \cxmark & 0.955 & 0.103 & 0.055 & 0.924 & 0.136 & 0.863  \\
       \ccmark & \cxmark & \ccmark & 0.958 & 0.099 & 0.051 & 0.924 & 0.128 & -  \\
       \rowcolor{gray!20} \ccmark & \ccmark & \ccmark & \textbf{0.958} & \textbf{0.099} & \textbf{0.050} & \textbf{0.928} & \textbf{0.123} & \textbf{0.879} \\
       \bottomrule
     \end{NiceTabular}%
   }
  }
\caption{Ablation experiment on loss selection.}
  \label{tab:ablation_loss}
\end{table}


\noindent \textbf{Comparison of Different Encoders.}
We conduct an ablation on the encoder backbone, comparing CLIP~\cite{radford2021learning}, SigLIP 2~\cite{tschannen2025siglip}, MetaCLIP 2~\cite{chuang2025meta} and DINOv3~\cite{simeoni2025dinov3} (with dino.txt~\cite{jose2025dinov2}). In all experiments, the encoders were frozen, with only the projection layer, transformer detector, and decoder heads being fine-tuned. As shown in Table~\ref{tab:ablation_encoder}, the DINOv3-based model achieves the best performance on nearly all metrics, highlighting its superior visual-text alignment and understanding for PGE.

\begin{table}[t]
  \centering
  {\fontsize{8pt}{11pt}\selectfont
   \resizebox{1.00\linewidth}{!}{%
     \begin{NiceTabular}{l l c c c c c c}[colortbl-like]
       \toprule[1.2pt]
       \multirow{2}{*}{\textbf{Encoder}} & \multirow{2}{*}{\textbf{Param}} & \multicolumn{3}{c}{\textbf{GazeFollow-Concept}} & \multicolumn{3}{c}{\textbf{VAT-Concept}} \\ \cmidrule(lr){3-5} \cmidrule(lr){6-8}
       & & AUC \(\uparrow\) & Avg L2 \(\downarrow\) & Min L2 \(\downarrow\) & AUC \(\uparrow\) & L2 \(\downarrow\) & AP \(\uparrow\)  \\
       \midrule[1.2pt]
       CLIP-B~\cite{radford2021learning} & 152M & 0.942 & 0.123 & 0.070 &  0.908 & 0.145 & 0.837 \\
       CLIP-L~\cite{radford2021learning} & 430M & 0.953 & 0.105 & 0.056 & 0.913 & 0.137 & 0.874  \\
       SigLIP2-B~\cite{tschannen2025siglip} & 379M & 0.949 & 0.115 & 0.064 & 0.904 & 0.148 & 0.860 \\
       SigLIP2-L~\cite{tschannen2025siglip} & 886M & 0.953 & 0.105 & 0.056 & 0.910 & 0.147 & 0.858   \\
       DINOv3-L~\cite{simeoni2025dinov3,jose2025dinov2} & 870M & \textbf{0.958} & \textbf{0.099} & \textbf{0.050} & \textbf{0.928} & \textbf{0.123} & \textbf{0.879}  \\
       MetaCLIP2-H~\cite{chuang2025meta} & 1.9B & 0.951 & 0.109 & 0.059 & 0.912 & 0.150 & 0.857 \\
       \bottomrule
     \end{NiceTabular}%
   }
  }
\caption{Comparison of different encoders for GazeAnywhere.}
  \label{tab:ablation_encoder}
  \vspace{-2mm}
\end{table}

\subsection{Visualization}

\begin{figure*}[t]
    \centering
    \includegraphics[width=1.0\linewidth]{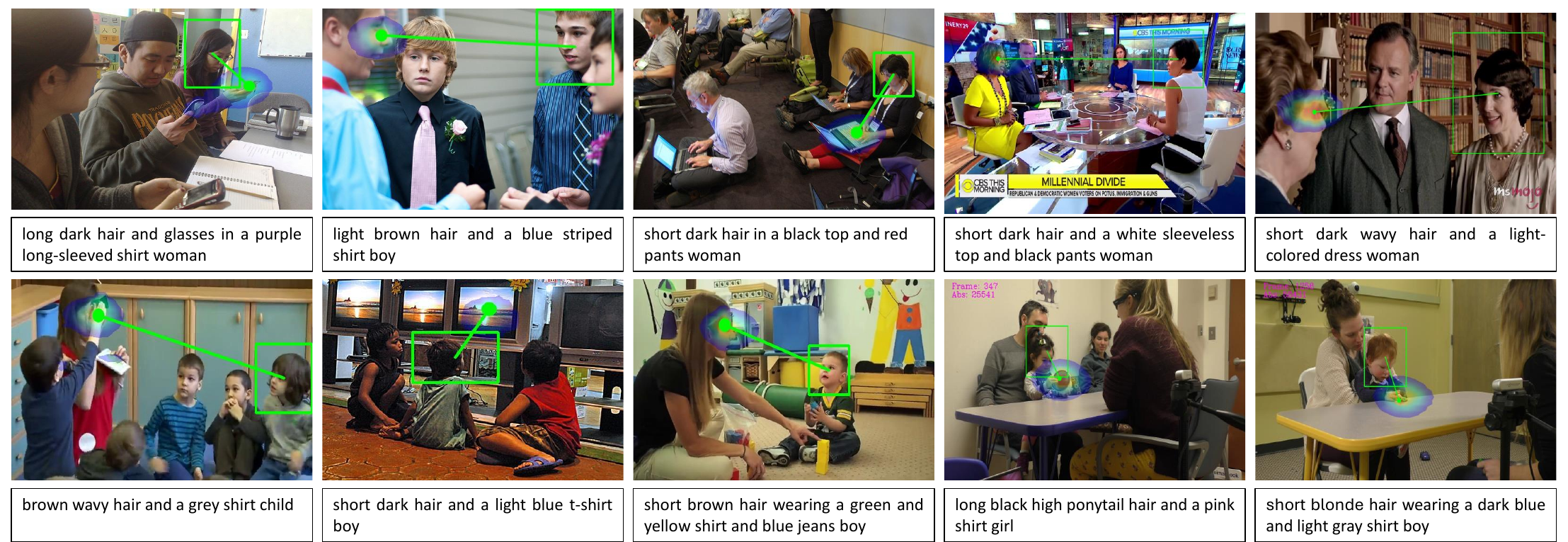}
    \caption{Visualization of GazeAnywhere's gaze target estimation results from several datasets. The input is the text prompt (shown in the black box) describing only the subject person's appearance and the image. GazeAnywhere detect the subject's head and track the gaze target. More qualitative comparison is shown in the appendix.}
    \label{fig:visualization}
    \vspace{-2mm}
\end{figure*}

Figure~\ref{fig:visualization} showcases qualitative gaze estimation results from GazeAnywhere. The input, displayed in the black boxes, is a text prompt describing only the subject's appearance, such as "light brown hair and a blue striped shirt boy". The visualizations demonstrate that GazeAnywhere performs robustly not only in simple scenarios with 2-3 people but also in complex, crowded scenes with four or more individuals. Notably, the final two examples, "long black high ponytail hair and a pink shirt girl" and "short blonde hair wearing a dark blue and light gray shirt boy", are from an OOD Child-SC video dataset. The model's successful performance on this unseen data highlights its generalization and robustness.

\subsection{GazeAnywhere as Agent in AR}

\begin{figure}[t]
    \centering
    \includegraphics[width=1.0\linewidth]{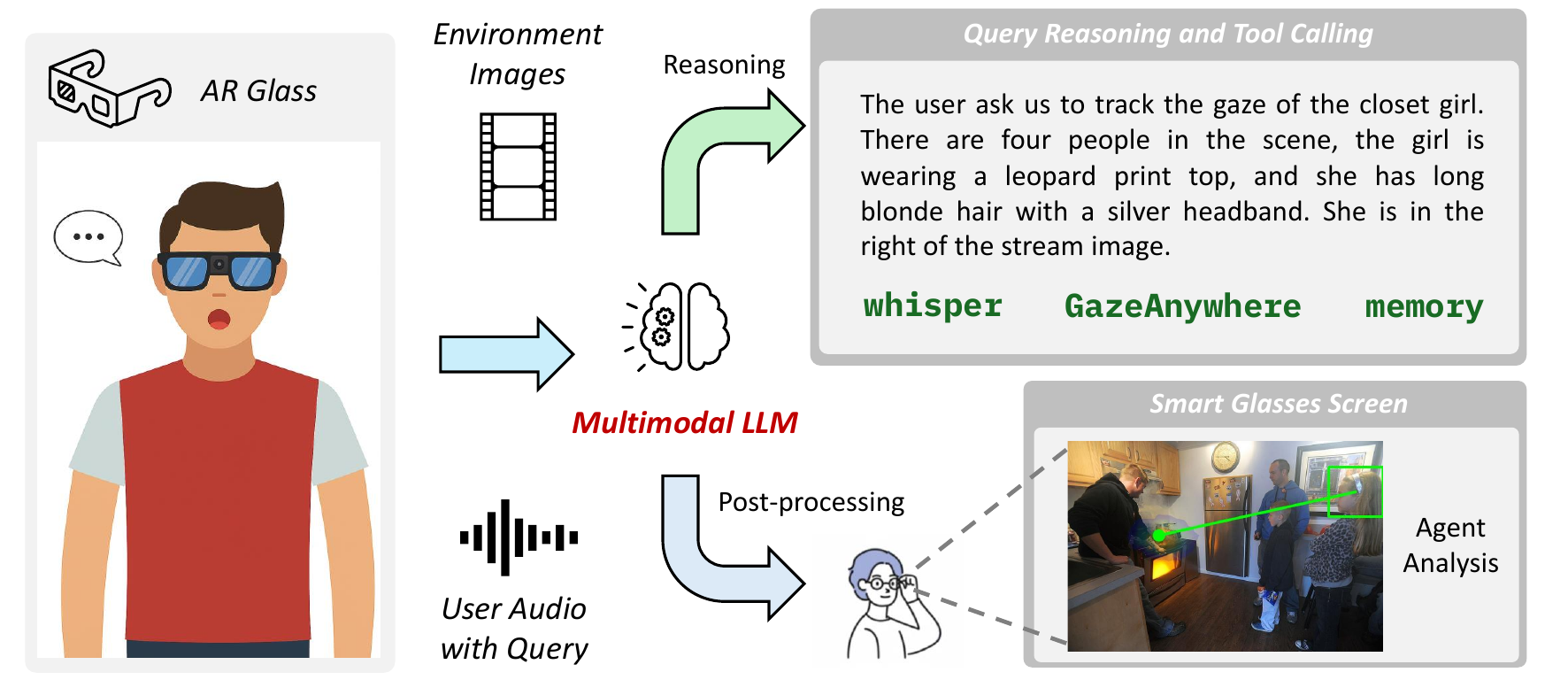}
    \caption{Workflow of MLLM-powered GazeAnywhere Agent.}
    \label{fig:agent}
    \vspace{-2mm}
\end{figure}

Previous gaze estimation models ignore the real-world application experiment. Inspired by recent tool-enhanced MLLM workflows~\cite{yang2025contextagent,carion2025sam}, we developed the GazeAnywhere Agent (Figure~\ref{fig:agent}). This system uses a central MLLM (Gemini 2.5~\cite{comanici2025gemini}) that leverages GazeAnywhere as a specialized tool to solve advanced user queries, such as, "How many gaze shifts does this girl with the white dress present?". The workflow captures User Audio and Environment Images from an AR Glass. MLLM calls Whisper v3~\cite{radford2023robust} to transcribe the audio, followed by query reasoning and prompt rephrasing. The agent converts the high-level query into a low-level text prompt (e.g., "girl with white dress") , calls the GazeAnywhere tool to generate gaze tracing, and then uses its VLM function to analyze the post-processed video, providing the user with the required analysis.

We collected 10 real-world videos with rich gaze movement using the DigiLens ARGO AR glass to test the agent's performance. The evaluation focused on two tasks: gaze shift calculation and eye contact calculation with other social partners. Using the Mean Absolute Error (MAE) per-minute, the GazeAnywhere Agent demonstrated significantly better performance than a raw, single MLLM solution. Results of GazeAnywhere Agent experiment is showed in Table~\ref{tab:deployment_agent}.

\begin{table}[t]
  \centering
  {\fontsize{8pt}{11pt}\selectfont
   \resizebox{0.95\linewidth}{!}{%
     \begin{NiceTabular}{l c c c}[colortbl-like]
       \toprule[1.2pt]
       \textbf{MLLM}  & \textbf{Gaze Shift MAE/min \(\downarrow\)} & \textbf{Eye Contact MAE/min \(\downarrow\)} \\
       \midrule[1.2pt]
        Gemini 2.5 Flash (Raw) & 9.247 & 14.763\\ 
        Gemini 2.5 Flash (AnyGaze Agent) & 2.337& 4.756\\
        Gemini 2.5 Pro (Raw) & 6.063 & 10.268\\
        Gemini 2.5 Pro (AnyGaze Agent) & 2.546 & 6.672\\
       \bottomrule
     \end{NiceTabular}%
   }
  }
\caption{Comparison of GazeAnywhere Agent and no agent framework.}
  \label{tab:deployment_agent}
  \vspace{-4mm}
\end{table}

\section{Discussion}

The practical applications of human gaze target estimation are diverse and impactful. In healthcare, for instance, this technology can significantly enhance the analysis of non-verbal communication behaviors which are implicated in the diagnosis and treatment of developmental conditions such as autism~\cite{rehg2014behavioral}.
In order for AI models to be used in clinical applications, they must be sufficiently robust and easy to use by nonexperts. This works takes a significant step in that direction for the task of gaze assessment. Our concept-based approach, which allows subjects to be identified by their attributes in natural language, is a first step towards the flexible and convenient specification of a broad set of behavioral analysis tasks. In addition, by creating a unified end-to-end learnable architecture we increase robustness by eliminating brittle stage-wise approaches to identifying the subjects of gaze analysis. Our approach is beneficial even in comparison to using state-of-the-art OVD models to identify subjects, e.g. the SOTA OVD OWLv2 has only a 70\% detection accurate rate in Child-SC for the child head and face detection tasks. 


\section{Conclusion}
We present GazeAnywhere, a system that enables interactive human gaze target estimation using flexible, open-vocabulary text prompts to identify the subject. Our principal contributions include introducing the novel Promptable Gaze Target Estimation (PGE) task and Gaze-Co benchmark, proposing a tailored transformer-based detector and learning objective, and developing a human-and-AI-in-the-loop data engine to adapt existing datasets. GazeAnywhere achieves state-of-the-art results in Gaze-Co benchmark, and its robustness is further validated on a challenging out-of-domain (OOD) dataset of child social communication videos. We believe GazeAnywhere and the Gaze-Co benchmark represent important milestones, paving the way for future research and applications in social AI and human behavior understanding.

\clearpage

\section*{Acknowledgments}
Portions of this work were supported in part by NIH R01 MH114999, the CIFAR Child and Brain Development program, and the Health Care Engineering Systems Center at University of Illinois Urbana-Champaign. Gemini API used in the project is supported by Google. This work also used Delta at the National Center for Supercomputing Applications (NCSA) through allocation CIS251391 from the Advanced Cyberinfrastructure Coordination Ecosystem: Services \& Support (ACCESS) program~\cite{boerner2023access}, which is supported by U.S. National Science Foundation grants \#2138259, \#2138286, \#2138307, \#2137603, and \#2138296.  

{
    \small
    \bibliographystyle{ieeenat_fullname}
    \bibliography{main}
}

\clearpage

\clearpage
\setcounter{page}{1}
\maketitlesupplementary

\section{Further Discussions \& Social Impact}

\subsection{Toward End-to-end Gaze Target Estimation}

The evolution of human gaze estimation shows a clear trend: a move away from complex auxiliary features like pose and depth towards streamlined, head box-only inputs~\cite{chong2018connecting,fan2018inferring,fang2021dual,gupta2022modular,kellnhofer2019gaze360,lian2018believe,marin2019laeo,chen2021gaze,doosti2021boosting,zhao2020learning}. This simplification has spurred the development of end-to-end, OpenPose~\cite{cao2017realtime}-like, and DETR~\cite{carion2020end}-like bottom-up approaches that can detect all head box-gaze pairs within a scene~\cite{tonini2022multimodal,tonini2023object,tu2022end,de2025gazedetr}. However, a critical limitation persists. These methods lack identity association; they can find the gaze of everyone but cannot identify the gaze of a specific person. This necessitates separate modules or post-processing to link a detected gaze to a particular individual. Thus, the cascaded detection error will still exist. Our work, GazeAnywhere, directly addresses this gap. We propose the first text-promptable pipeline that simultaneously resolves human identification and gaze target estimation, enabling targeted queries for a specific person's gaze.

\subsection{Future Application}

Joint attention, the capability of following another person's head turn and gaze direction, typically emerges in children with Autism Spectrum Disorder (ASD) years later than in typically developing children~\cite{leekam1998targets}. Previous research has demonstrated the strong potential of gaze target estimation models to capture these atypical joint attention behaviors, offering a promising avenue for the early screening and detection of ASD~\cite{marin2011here,chong2017detecting}. The concept-based prompting flexibility of the GazeAnywhere model offers a significant evolution in this domain. In future clinical and home-based settings, this model could be deployed to continuously and non-invasively track a child's gaze behavior. Critically, GazeAnywhere and GazeAnywhere Agent can be used by pediatricians or even the patient's parents simply by describing the patient's appearance or location in the prompt (e.g., "the child in the blue shirt"), thereby omitting the complicated and labor-intensive process of manually drawing head bounding boxes for annotation. This simplified usage offers the benefit of longitudinal tracking outside of a clinical setting, enabling earlier intervention and more comprehensive developmental monitoring. In addition, user can query GazeAnywhere Agent to let MLLM post-process the target tracking video and provide high-level gaze behavior information like gaze shift.

\section{GazeAnywhere Agent}

\begin{figure*}[t]
    \centering
    \includegraphics[width=1.0\linewidth]{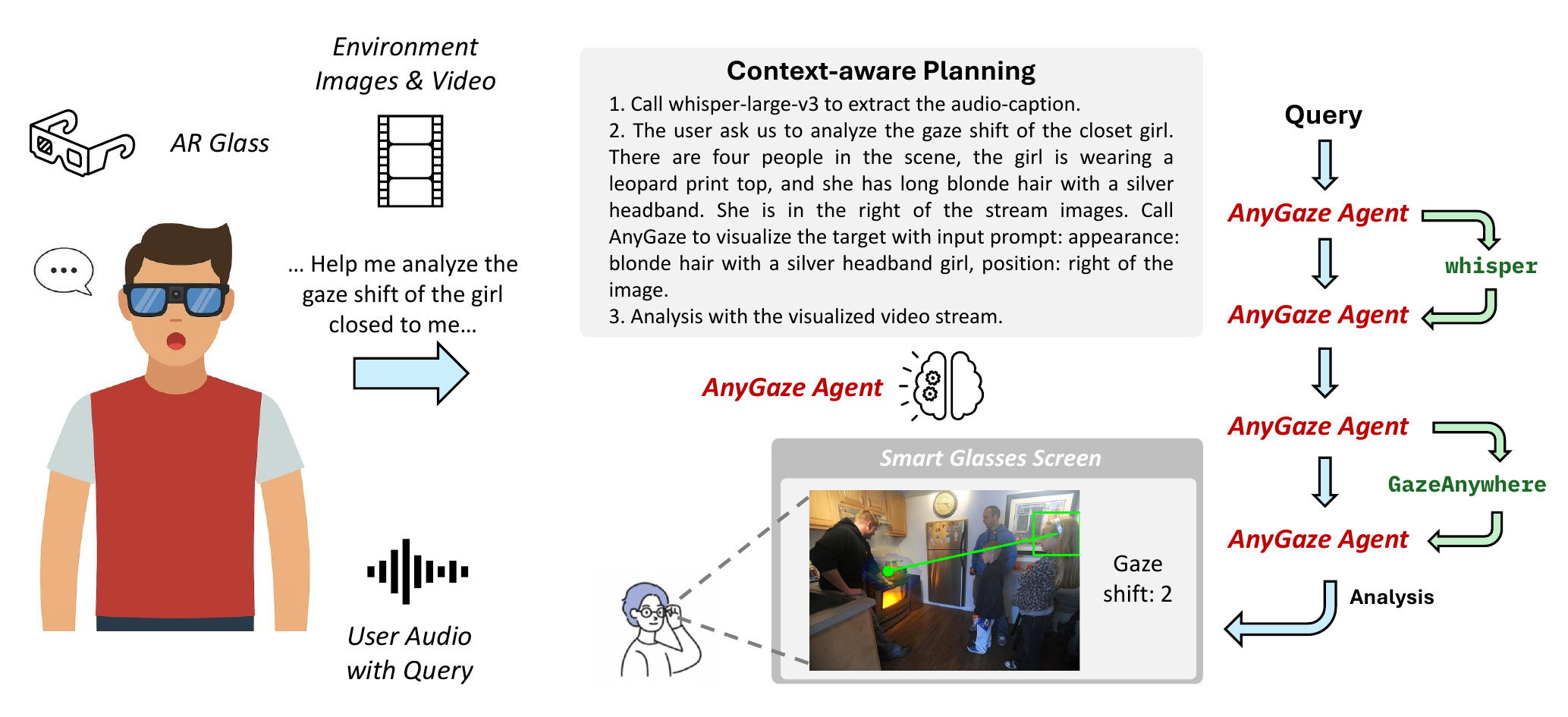}
    \caption{Step-by-step explanation of how GazeAnywhere Agent works.}
    \label{fig:agent_detail}
\end{figure*}

In this section, we introduce the GazeAnywhere Agent, a visual agentic framework designed to process natural-language gaze estimation and post-analysis requests. Figure~\ref{fig:agent_detail} illustrate the workflow of the agent. The system dynamically queries a MLLM to orchestrate specific tools. The initial version of the agent integrates two primary models as the tool: Whisper-large-v3 for audio-to-text conversion and our proposed GazeAnywhere model for PGE target prediction.

Given an input image or video and a user request via audio , the MLLM acts as a planner and controller. It first converts the user's audio to text, analyzes the scene context, devises a step-by-step plan, and subsequently invokes the GazeAnywhere model. After each action, the agent receives visual feedback by visualizing the gaze target within the scene. This feedback is stored in memory, enabling the agent to revise its plan and determine the next steps for analysis. This pipeline handles queries far more complex than simple noun phrases, facilitating a deeper understanding of human gaze behavior in video streams.

\section{Dataset \& Benchmark}

\subsection{Training Set}
\begin{figure*}[t]
    \centering
    \newcommand{\trainPieHeight}{0.20\textwidth}
    \subfloat[Source dataset distribution]{%
        \includegraphics[height=\trainPieHeight]{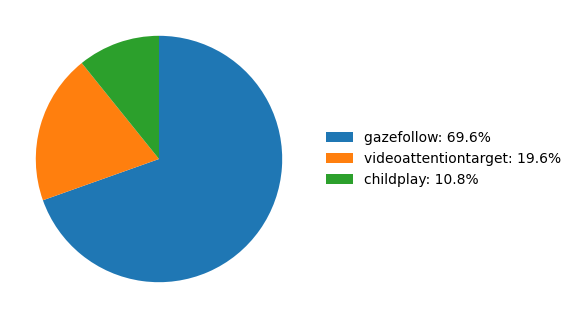}%
        \label{fig:dist_a}
    }\hfill
    \subfloat[Apparent subject category distribution]{%
        \includegraphics[height=\trainPieHeight]{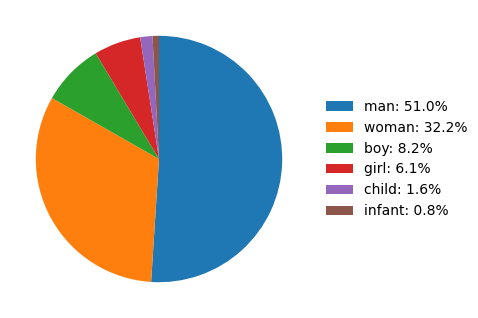}%
        \label{fig:dist_b}
    }\hfill
    \subfloat[In-frame vs out-of-frame gaze distribution]{%
        \includegraphics[height=\trainPieHeight]{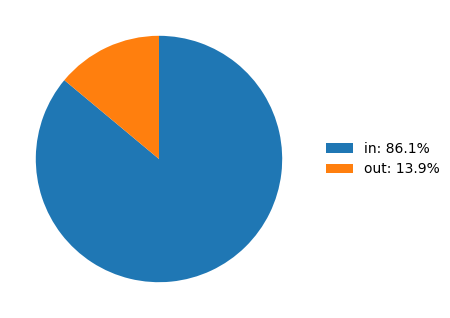}%
        \label{fig:dist_c}
    }
    \caption{Training-set statistics of the Gaze-Co dataset: (a) proportion of each source dataset, (b) distribution of apparent subject categories, and (c) proportion of in-frame vs out-of-frame gaze annotations.}
    \label{fig:dist_all}
\end{figure*}

The Gaze-Co training set contains 119{,}525 samples in total. Each record includes the target head bounding box, normalized gaze point, an in/out-of-frame label, and a compact concept phrase (attribute, position, action, and pose). The training data are constructed from three published gaze datasets after applying the image-quality filters, MLLM-based concept generation, and human in-loop MLLM verification described in the main text. In terms of source datasets, 69.6\% (83{,}148 samples) come from GazeFollow~\cite{recasens2015they,recasens2017following}, 19.6\% (23{,}481 samples) from  VideoAttentionTarget~\cite{chong2020detecting}, and 10.8\% (12{,}896 samples) from ChildPlay~\cite{tafasca2023childplay} (see Fig.~\ref{fig:dist_a}).

For the apparent subject category, 51.0\% (60{,}983 samples) are labeled as man, 32.2\% (38{,}508) as woman, 8.2\% (9{,}773) as boy, 6.1\% (7{,}337) as girl, 1.6\% (1{,}916) as child (unspecified gender), and 0.8\% (1{,}008) as infant (unspecified gender) (Fig.~\ref{fig:dist_b}). These labels reflect perceived visual categories rather than verified identity attributes. Regarding gaze location, 13.9\% (16{,}671 samples) of annotations are out-of-frame, while 86.1\% (102{,}854) fall within the image (Fig.~\ref{fig:dist_c}).

\subsection{Concept-based In-domain Test Set}
We derive three concept-augmented test splits by converting the official test splits of GazeFollow, VAT, and ChildPlay into our unified PGE schema (image, head box, normalized gaze point, in/out-of-frame label, and concept phrase). After applying the same image-quality filters as in the training set, we obtain GazeFollow-Concept, VAT-Concept, and ChildPlay-Concept. To guarantee a high quality benchmark for both baselines and our model evaluation, all concept annotations are human verified instead of using MLLM.

 \noindent \textbf{GazeFollow-Concept.}
After filtering, GazeFollow-Concept contains 2{,}436 (image, head box) records. In terms of apparent subject category, 49.1\% (1{,}197 samples) are labeled as man, 31.7\% (772) as woman, 10.3\% (250) as boy, 5.6\% (136) as girl, 2.1\% (50) as child (unspecified gender), and 1.3\% (31) as infant (unspecified gender). All annotations in this split correspond to in-frame gaze targets (100\%, 2{,}436 samples). In the dataset, each (image, head box) record is associated with multiple human gaze point annotations from the original GazeFollow dataset, which motivates the additional Avg L2 and Min L2 metrics used in the main text: Avg L2 is defined as the distance between the predicted gaze point and the mean of all human annotations, and Min L2 as the distance to the nearest human-annotated gaze point.

\noindent  \textbf{VAT-Concept.}
VAT-Concept contains 5{,}301 records. For apparent subject categories, 45.9\% (2{,}435 samples) are labeled as man, 43.8\% (2{,}324) as woman, 5.8\% (310) as boy, 0.4\% (20) as girl, 0.1\% (5) as child (unspecified gender), and 3.9\% (207) as infant (unspecified gender). Regarding gaze location, 35.5\% (1{,}884 samples) of annotations are out-of-frame, while 64.5\% (3{,}417) are in-frame.

\noindent  \textbf{ChildPlay-Concept.}
ChildPlay-Concept contains 1{,}238 records. In terms of apparent subject category, 8.5\% (105 samples) are labeled as man, 32.9\% (407) as woman, 36.8\% (455) as boy, 12.0\% (148) as girl, 4.4\% (55) as child (unspecified gender), and 5.5\% (68) as infant (unspecified gender). For gaze location, 14.7\% (182 samples) of annotations are out-of-frame, while 85.3\% (1{,}056) are in-frame.

\noindent
Across all splits, the apparent subject categories reflect perceived visual attributes rather than verified identity labels.

\subsection{Concept-based Out-of-domain Test set}
For out-of-domain evaluation, we utilize Child–Social Communication (Child-SC), a private dataset protected by IRB. It captures natural interactions between children and clinicians, where the clinician guides the child’s attention across various targets using toys, thus eliciting frequent and structured gaze shifts. The dataset comprises 326 video clips from 40 children, sampled at 5 fps, yielding a total of 151,533 images. Due to privacy regulations, these images cannot be processed by cloud-based MLLM; consequently, all target-person concepts were manually annotated, strictly adhering to the style and protocols of our MLLM-generated concepts.

\section{Baseline Details}

\subsection{Open-Vocabulary Detector (OVD)}
As baselines, we use the OVD models to locate the target person described by a text prompt. This step supports our main task: to predict the point of view of the subject. Each OVD model takes an image and a prompt, matches text to visual regions in a shared vision–language space, and scores candidate boxes by text–image similarity. It outputs the highest-confidence bounding box for the prompted person, which we use as the subject-person localization. We also compared with the SOTA open-vocabulary human detection model RexSeek~\cite{jiang2025referring}, which is a 3B foundation model in referring expression comprehension task.


\paragraph{GroundingDINO-B.} GroundingDINO-B~\cite{liu2024grounding} is a Transformer-based detector featuring a dual-encoder single-decoder architecture that deeply fuses image and text features. It employs a language-guided query selection module to initialize object queries based on the input prompt. This mechanism produces a series of refined candidate boxes associated with prediction scores. From these outputs, we identify the target person by selecting the box with the highest confidence score for the referring phrase.

\paragraph{LLMDet-L.} LLMDet-L~\cite{fu2025llmdet} enhances open-vocabulary detection through multimodal co-training, where a large language model generates detailed captions to enrich feature alignment during training. At test time, with the LLM removed, the detector takes the image and prompt to generate multiple region candidates. It evaluates these regions by matching them against the text embedding, enabling us to filter the results and retrieve the top-ranked bounding box as the localized subject.

\paragraph{OWLv2-L.} OWLv2-L~\cite{minderer2023scaling} scales up the OWL-ViT architecture using a massive self-training strategy on over one billion weakly supervised examples. It utilizes a Vision Transformer backbone to directly predict bounding boxes and text-alignment scores from image tokens. When queried with the target person's description, the model outputs a collection of detected objects with their semantic similarity scores, from which we select the best-matching candidate to localize the person.

\subsection{Gaze Model}
Following the localization step, we evaluate several gaze-following models to predict the target person's point of regard. These models accept the full scene image and the localized person region as input. They output a 2D gaze heatmap (probability distribution), and we extract the coordinates of the peak value from the heatmap to represent the final predicted gaze location.

\paragraph{ViTGaze}
ViTGaze~\cite{song2024vitgaze} is a single-modality gaze-following model that predicts a person’s gaze target using RGB information only. Given the full image and the target person’s head bounding box, it employs a pre-trained ViT to extract human–scene interaction cues directly from self-attention maps, eliminating the need for extra modalities. The model outputs a 2D gaze heatmap along with an in/out-of-frame score for evaluation.

\paragraph{Sharingan}
Sharingan~\cite{tafasca2024sharingan} introduces a transformer-based architecture designed to capture global gaze interactions. It represents the target person via a Person Gaze Token, constructed by fusing head-crop features with normalized head-box coordinates. This token is processed with scene tokens by a ViT encoder to model human–scene dependencies. The model outputs a 2D gaze heatmap representing the spatial probability of the gaze target and an in/out-of-frame score. 

\paragraph{Gaze-LLE}
Gaze-LLE~\cite{ryan2025gaze}
is a streamlined estimator built on a frozen, large-scale DINOv2 encoder, departing from traditional multi-branch head/scene architectures. Given the full image and the target person’s head bounding box, it encodes the head location as a positional prompt injected into the scene features, using a lightweight transformer decoder to model head–scene relations. The model predicts a 2D gaze heatmap along with an in-/out-of-frame score.
\section{Experimental Protocol}

\subsection{AR Device for GazeAnywhere Agent}

\begin{figure}[t]
    \centering
    \includegraphics[width=1.0\linewidth]{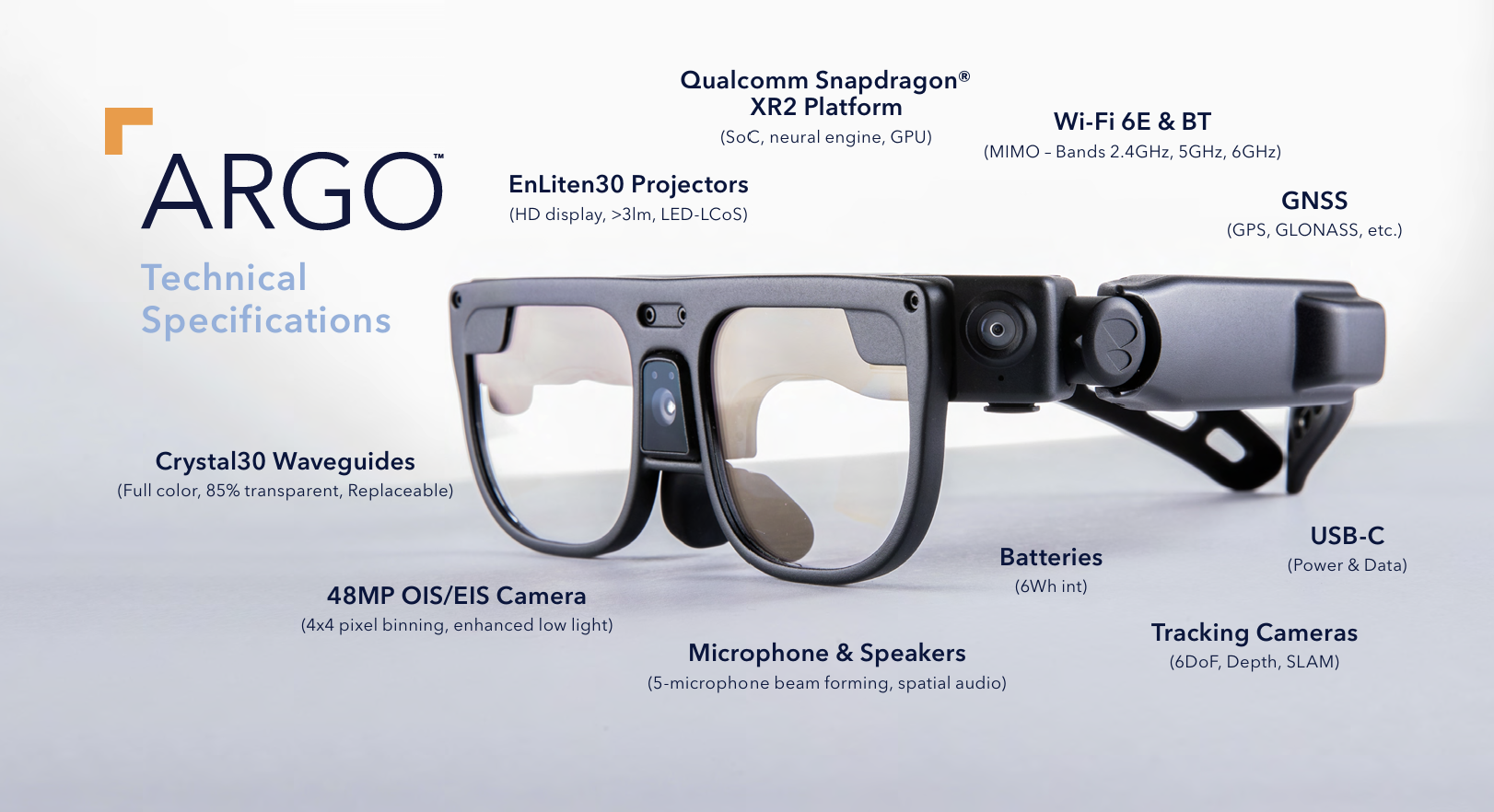}
    \caption{DigiLens ARGO AR glasses used for video and audio capture and on-device feedback in the GazeAnywhere Agent.}
    \label{fig:glass}
\end{figure}

We use DigiLens ARGO in the experiment to capture video data in real-world settings (Fig.~\ref{fig:glass}). Its 48 MP camera records high-resolution video with autofocus, optical and electronic stabilization, 4×4 pixel binning, and strong low-light support. For audio, a five-microphone beamforming array is designed to pick up the wearer’s voice in noisy environments and provides spatial recordings suitable for analysis.

\subsection{Implementation Details of GazeAnywhere-DINOv3-L}

The deployed version of GazeAnywhere-DINOv3-L consists of a detector transformer with 3 layers and a dimension of $D=256$. Both the visual and text prompts are trained jointly. For visual prompting, we apply diverse augmentation techniques during training, including head/body bounding box jittering, color jittering, random resizing and cropping, horizontal flipping, rotation, and masking of scene patches. For text prompting, as the subject position text information is fixed, we limit visual augmentation to random scene patch masking and apply text augmentation with reordering appearance, location, pose, and action attributes. During training, the input resolution is $512\times512$.

\section{More Results}
\label{appendix:more_results}

\subsection{Impact of Frozen Encoder.} A key design choice for GazeAnywhere is to keep the image and text encoders frozen. We validate this approach in Table~\ref{tab:ablation_freeze}, which compares the default frozen model against one where the DINOv3 image encoder or the text encoder are fine-tuned. Unfreezing image or text encoders leads to a clear drop in performance. This demonstrates that DINOv3's pre-trained features are highly robust and generalizable for the PGE task, and that fine-tuning may lead to overfitting or harmful feature drift.

\begin{table}[t]
  \centering
  {\fontsize{8pt}{11pt}\selectfont
   \resizebox{1.00\linewidth}{!}{%
     \begin{NiceTabular}{l l c c c c c c c c}[colortbl-like]
       \toprule[1.2pt]
       \multirow{2}{*}{\textbf{Visual}} & \multirow{2}{*}{\textbf{Text}} &  \multirow{2}{*}{\textbf{Trainable Param}} & \multicolumn{3}{c}{\textbf{GazeFollow-Concept}} & \multicolumn{3}{c}{\textbf{VAT-Concept}} \\ \cmidrule(lr){4-6} \cmidrule(lr){7-9}
       & & & AUC \(\uparrow\) & Avg L2 \(\downarrow\) & Min L2 \(\downarrow\) & AUC \(\uparrow\) & L2 \(\downarrow\) & AP \(\uparrow\) \\
       \midrule[1.2pt]
       \cxmark & \cxmark & 870.2M & 0.931 & 0.150 & 0.095 & 0.886 & 0.212 & 0.823\\
       \ccmark & \cxmark & 332.1M & 0.943 & 0.119 & 0.067 & 0.886 & 0.206 & 0.813 \\
       \cxmark & \ccmark & 541.7M & 0.952 & 0.130 & 0.078 & 0.874 & 0.201 &  0.825 \\
       \rowcolor{gray!20} \ccmark & \ccmark & 3.6M & \textbf{0.958} & \textbf{0.099} & \textbf{0.050} & \textbf{0.928} & \textbf{0.123} & \textbf{0.879}  \\
       \bottomrule
     \end{NiceTabular}%
   }
  }
\caption{Comparison of the encoder frozen strategies.}
  \label{tab:ablation_freeze}
  \vspace{-1mm}
\end{table}

\subsection{Impact of Detector Dimension.} We study the impact of the Detector transformer's layer dimension $D$ in Table~\ref{tab:ablation_k}. The results indicate that performance plateaus at $D=128$. We observed no significant performance gain from increasing $D$ further, and thus selected $D=256$ as it provides the best trade-off between accuracy and computational cost.

\begin{table}[t]
  \centering
  {\fontsize{8pt}{11pt}\selectfont
   \resizebox{1.00\linewidth}{!}{%
     \begin{NiceTabular}{l c c c c c c c}[colortbl-like]
       \toprule[1.2pt]
       \multirow{2}{*}{\textbf{$D$ of $\psi(\cdot)$}} & \multicolumn{3}{c}{\textbf{GazeFollow-Concept}} & \multicolumn{3}{c}{\textbf{VAT-Concept}} \\ \cmidrule(lr){2-4} \cmidrule(lr){5-7}
       & AUC \(\uparrow\) & Avg L2 \(\downarrow\) & Min L2 \(\downarrow\) & AUC \(\uparrow\) & L2 \(\downarrow\) & AP \(\uparrow\) \\
       \midrule[1.2pt]
       64 & 0.953 & 0.115 & 0.062 & 0.915 & 0.144 & 0.871 \\
       128 & 0.960 & 0.100 & 0.050 & 0.928 & 0.116 & 0.875 \\
       256 & 0.958 & 0.099 & 0.050 & 0.928 & 0.123 & 0.879 \\
       512 & 0.959 & 0.104 & 0.054 & 0.917 & 0.122 & 0.875  \\
       \bottomrule
     \end{NiceTabular}%
   }
  }
\caption{Ablation experiment on the selection of the Detector Transformer dimension.}
  \label{tab:ablation_k}
\end{table}

\subsection{Ablation on Detector's Transformer Layer Number}

We conduct another ablation study to explore the layer number of transformer blocks in detector transformers. Results are shown in Table~\ref{tab:ablation_layer}. After increasing the layer number to 3, the model shows stable performance.

\begin{table}[t]
  \centering
  {\fontsize{8pt}{11pt}\selectfont
   \resizebox{0.95\linewidth}{!}{%
     \begin{NiceTabular}{l c c c c c c c}[colortbl-like]
       \toprule[1.2pt]
       \multirow{2}{*}{\textbf{Layer Num of $\psi(\cdot)$}} & \multicolumn{3}{c}{\textbf{GazeFollow-Concept}} & \multicolumn{3}{c}{\textbf{VAT-Concept}} \\ \cmidrule(lr){2-4} \cmidrule(lr){5-7}
       & AUC \(\uparrow\) & Avg L2 \(\downarrow\) & Min L2 \(\downarrow\) & AUC \(\uparrow\) & L2 \(\downarrow\) & AP \(\uparrow\) \\
       \midrule[1.2pt]
       1 & 0.949 & 0.126 & 0.076 & 0.882  & 0.179 & 0.846 \\
       2 & 0.957 & 0.104 & 0.054 & 0.920 & 0.123 & 0.871  \\
       3 & 0.958 & 0.099 & 0.050 & 0.928 & 0.123 & 0.879\\
       4 & 0.957 & 0.100 & 0.051 & 0.928 & 0.120 &  0.858 \\
       5 & 0.959 & 0.095 & 0.047 & 0.929 & 0.121 & 0.888 \\
       \bottomrule
     \end{NiceTabular}%
   }
  }
\caption{Ablation experiment on the selection of the transformer layer number in the Detector.}
  \label{tab:ablation_layer}
\end{table}

\section{Qualitative Analysis}

\begin{figure*}[t]
    \centering
    \includegraphics[width=\textwidth]{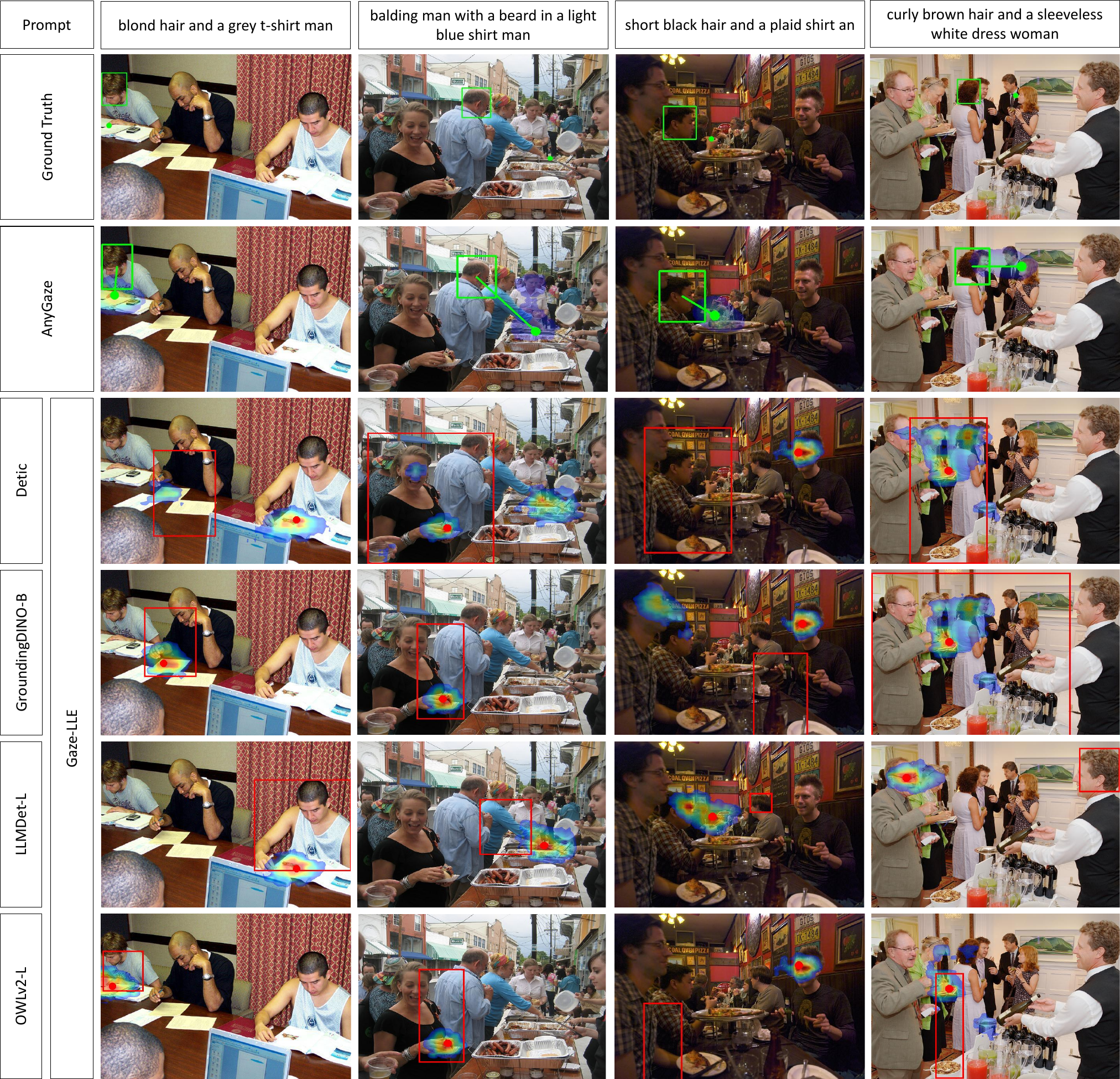}
    \caption{
        Qualitative comparison of gaze-target localization conditioned on appearance prompts.
        Each column corresponds to a different sample, and each row shows predictions
        from a different method. Our method produces sharper and more accurate heatmaps around
        the true gaze targets.
    }
    \label{fig:qualitative_comparison}
\end{figure*}




In Figure~\ref{fig:qualitative_comparison}, we qualitatively compare GazeAnywhere with the current state-of-the-art model, Gaze-LLE. Although Gaze-LLE performs well in sparse scenes with only one or two individuals, its performance degrades noticeably as crowd density increases. As shown in Figure~\ref{fig:qualitative_comparison}, the upstream OVD module becomes unreliable in these complex settings and typically fails in two ways. First, it may localize the wrong person, causing Gaze-LLE to estimate gaze for an incorrect target. Second, it may produce an overly large bounding box that covers multiple people; even if the true target is included, Gaze-LLE cannot reliably disambiguate whom to condition on. These examples expose a key limitation of two-stage gaze estimation pipelines in real-world social scenes.


\section{Related Prompts}
For reproducibility, we include the exact natural-language prompts used to query the MLLM in our pipeline.
These prompts support three major components: the concept-generation data engine, the MLLM-only gaze prediction baseline, and the GazeAnywhere Agent for video-based social gaze analysis.
Unless otherwise noted, the prompts are shown verbatim as used in our batch API calls.

\subsection{Data Engine}
This section summarizes the prompts used by the data engine to construct concept-level annotations for each subject person.
The attribute prompt (Fig.~\ref{fig:attribute_prompt}) instructs the MLLM to produce a compact description of appearance, position, action, pose, and people count for the person marked by the green head box.

The concept verification prompt (Fig.~\ref{fig:concept_verification_prompt}) then asks the MLLM to check, field by field, whether a candidate concept matches the image and to return JSON flags for attribute, position, action, pose, and an overall pass/fail decision.
Together with spot-checks from human annotators, these prompts implement the MLLM component of our human-in-the-loop data engine.
\begin{figure*}[t!]
\centering
\begin{planbox}{Concept Generation Prompt}
\small\ttfamily
TASK\\
Return a description for the person with a green bounding box in head:\\
The description is a natural, concise attribute phrase (<30 words in total).\\[0.5em]

STYLE \& CONTENT\\
- all lowercase\\
- avoid generic words: person, people, adult; avoid starting with a/an/the\\
- prefer stable visible attributes, in this order: hair style and color / hat > glasses / beard > top garment color and pattern > pant / dress garment color and pattern.\\
- the LAST word of attributes MUST be one of: man, woman, boy, girl, infant, child\\
- prefer short and clear location description, such as bottom left corner.\\
- action: describe ongoing interaction or movement; make it specific by adding target/object/direction when visible. keep “what is being done” here. if unclear, write "none".\\
- pose: describe static body configuration and facing direction; keep “how the body is” here (orientation, posture, limb arrangement). if unclear, write "none".\\
- keep action and pose distinct and non-overlapping.\\
- also provide an approximate count of people visible in the scene; report a single integer when feasible; if indeterminate, write "none".\\[0.5em]

You output format is:\\
<attribute> attributes of the human </attribute>\\
<position> position of the human in the camera </position>\\
<action> action of the human </action>\\
<pose> pose of the human </pose>\\
<count> estimated number of people in the scene </count>\\
\end{planbox}
\caption{\textbf{Concept generation prompt} used for generating concept phrases for the target person.}
\label{fig:attribute_prompt}
\end{figure*}
\begin{figure*}[t!]
\centering
\begin{planbox}{Concept Verification Prompt}
\small\ttfamily
TASK\\
You see an image with one green bounding box on a human head and a candidate description from Prompt A, with:\\
<attribute>, <position>, <action>, <pose>, <count>.\\
Check whether the first four fields match the target human and the scene. Ignore <count>.\\[0.5em]

CHECKING RULES\\[0.25em]

attribute\\
must describe the same person in the green box.\\
hair / hat / glasses / beard / clothes must match.\\
last word must be one of: man, woman, boy, girl, infant, child.\\
label as correct only if all above are satisfied.\\[0.5em]

position\\
must match the boxed human location (e.g., top left, bottom center, center right).\\
label as correct only if consistent with boxed human position.\\[0.5em]

action\\
must be a visible ongoing movement or interaction of this person.\\
if not clearly visible, the correct value should be "none".\\
label as correct only if supported by the image.\\[0.5em]

pose\\
must describe static body configuration and facing direction (orientation, posture, limb arrangement).\\
must be distinct from action; if unclear, should be "none".\\
label as correct only if supported by the image and distinct from action.\\[0.5em]

OVERALL\\
overall is "pass" only if all four checks are "correct".\\
otherwise overall is "fail".\\[0.5em]

OUTPUT FORMAT\\
Output only a single JSON object with exactly these keys and values:\\[0.25em]

"attribute\_check": "correct" or "incorrect"\\
"position\_check": "correct" or "incorrect"\\
"action\_check": "correct" or "incorrect"\\
"pose\_check": "correct" or "incorrect"\\
"overall": "pass" or "fail"\\
\end{planbox}
\caption{\textbf{Concept verification prompt} used to check attribute, position, action, and pose consistency for each subject.}
\label{fig:concept_verification_prompt}
\end{figure*}

\subsection{MLLM Baseline}
Here we provide the prompt used for the MLLM-only gaze target prediction baselines, Gemini-2.5-flash (Fig.~\ref{fig:gaze_target_prompt_gemini}) and Qwen3-VL-8b (Fig.~\ref{fig:gaze_target_prompt_qwen3vl}).

Given an image and a textual concept description, the model is asked to predict an gaze in/out-frame flag and a normalized 2D gaze target point, and to return the answer in a strict JSON format.

\begin{figure*}[t!]
\centering
\begin{planbox}{Gaze Target Prediction Prompt of Gemini 2.5 Flash}
\small\ttfamily
You are given an image, where the top-left corner is (0, 0) and the bottom-right corner is (1, 1).\\[0.5em]

Coordinates are normalized by the image width and height.\\
You are also given a description of the subject person in the image:\\[0.5em]

Attribute: \{attribute\}\\
Location: \{position\}\\
Action: \{action\}\\
Pose: \{pose\}\\
Based on this description and the image, perform the following tasks:\\[0.5em]

1. In-frame gaze flag\\
Indicate whether the subject person is looking at a target inside the image frame.\\[0.25em]

Output 1 if the gaze target lies within the image frame.\\
Output 0 if the subject person is looking outside the image frame.\\[0.5em]

2. Gaze target point\\
Predict the gaze target of the subject person as a point $(x,y) \in [0,1]$, with exactly three decimal places for both x and y.\\[0.25em]

The values must be normalized by the image width and height.\\[0.5em]

Output format\\
Return only a valid JSON object, with no extra text, in the following format:\\
\{\\
\ \ "in\_frame\_gaze": 0,\\
\ \ "gaze\_target": \{\\
\ \ \ \ "x": 0.XXX,\\
\ \ \ \ "y": 0.XXX\\
\ \ \}\\
\}\\
in\_frame\_gaze must be either 0 or 1.\\
x and y must be numbers in [0,1] with three decimal places.\\
\end{planbox}
\caption{\textbf{Gaze target prediction prompt} used for in-frame flagging and point estimation on Gemini-2.5 baseline}
\label{fig:gaze_target_prompt_gemini}
\end{figure*}

\begin{figure*}[t!]
\centering
\begin{planbox}{Gaze Target Prediction Prompt of Qwen3-VL.}
\small\ttfamily
You are given an image, where the top-left corner is (0.000 , 0.000), the bottom-right corner is (1.000, 1.000). The pixel point in the image is normalized to 0.000 to 1.000. All values are rounded by 3. \\[0.5em]
Here is the description of the subject person {Question} \\[0.5em]
Based on the description of a subject person in the image, perform the following task: \\[0.5em]
(1) Indicate whether the subject person is looking at a target inside the image frame. \\[0.5em]
Output 1 if the gaze target lies within the image frame. Output 0 if the subject person is looking outside the image frame. \\[0.5em]
Provide the inside prediction between the <inside> and </inside> tags. \\[0.5em]
(2) Predict the gaze target of the subject person as a point (x, y) x and y are in [0.000, 1.000]. If looking outside, randomly give values. \\[0.5em]
Provide the x of point between the <x> and </x> tags, y of point between the <y> and </y> tags. \\[0.5em]
\end{planbox}
\caption{\textbf{Gaze target prediction prompt} used for in-frame flagging and point estimation on Qwen3-VL-8B.}
\label{fig:gaze_target_prompt_qwen3vl}
\end{figure*}

\subsection{GazeAnywhere Agent}
This section lists the prompts used to compare gaze-target analysis with an MLLM alone versus an MLLM assisted by the GazeAnywhere Agent on smart-glasses recordings.
The raw-video prompt (Fig.~\ref{fig:gaze_shift_prompt_raw}) presents the model with the original AR recording and asks it to infer social gaze behavior directly from the unannotated video.

The GazeAnywhere-agent prompt (Fig.~\ref{fig:gaze_shift_prompt_agent}) uses the same video but with GazeAnywhere overlays (subject head box, gaze point, and out-of-frame indications), and instructs the model to count gaze shifts to social partners and overall gaze shifts.

\begin{figure*}[t!]
\centering
\begin{planbox}{Gaze Shift Analysis Prompt on Single MLLM Solution}
\small\ttfamily
This is a short video for human gaze target understanding. Can you give me an analysis of these tasks"?\\[0.5em]

The subject child is: short black hair white dress girl\\[0.5em]

1. "Gaze shift to social partner": The number of gaze shifts to the nearby person happened.\\[0.5em]

2. "Total gaze shift count": The number of gaze shifts happened. (change the gaze target to another object or out-of-frame)\\[0.5em]

Hint: Gaze shifting is the coordinated movement of the eyes and head to look at a new target. Per gaze shift means changing the gaze target from one object/human to another object/human\\[0.5em]

You should analyze the video and provide the answers to the above tasks.\\
\end{planbox}
\caption{\textbf{Gaze shift analysis prompt} used for counting gaze shifts and eye contact events on Single MLLM.}
\label{fig:gaze_shift_prompt_raw}
\end{figure*}

\begin{figure*}[t!]
\centering
\begin{planbox}{Gaze Shift Analysis Prompt on GazeAnywhere Agent}
\small\ttfamily
This is a short video for human gaze target understanding. The green bounding box is the detected subject child. If the bounding box's color becomes blue, it indicates the subject is looking out of the frame. The green point is the child's gaze target, and we overlap it with the raw video. Can you give me an analysis of these tasks"?\\[0.5em]

The subject child is: short black hair white dress girl\\[0.5em]

1. "Gaze shift to social partner": The number of gaze shifts to the nearby person happened.\\[0.5em]

2. "Total gaze shift count": The number of gaze shifts happened. (change the gaze target to another object or out-of-frame)\\[0.5em]

Hint: Gaze shifting is the coordinated movement of the eyes and head to look at a new target. Our green point in the frame can indicate the target location. So you should infer if the target is changed. Per gaze shift means changing the gaze target from one object/human to another object/human\\[0.5em]

You should analyze the video and provide the answers to the above tasks.\\
\end{planbox}
\caption{\textbf{Gaze shift analysis prompt} used for counting gaze shifts and eye contact events on GazeAnywhere agent.}
\label{fig:gaze_shift_prompt_agent}
\end{figure*}

\section{Notations}

We present the description all the notations in our paper in the last two pages.

\clearpage

\onecolumn
\begin{longtable}{p{4cm} p{12cm}}
    \hline
    \multicolumn{2}{l}{\textbf{Data and Indices}} \\
    \hline
$H$ & Height of input image \\
$W$ & Width of input image \\
$I \in \mathbb{R}^{3 \times H \times W}$ & Input RGB image \\
$P$ & Prompt \\
$T$ & Text \\
$H_{out}$ & Height of output image \\
$W_{out}$ & Width of output image \\
$\hat{H} \in \mathbb{R}^{H_{out} \times W_{out}}$ & Gaze heatmap \\
\midrule 
\multicolumn{2}{l}{\textbf{Embeddings and Image Encodings}} \\
\midrule
$\phi_V(\cdot)$ & Image encoder \\
$N_V$ & Number of patch tokens \\
$D_V$ & Visual embedding dimension \\
$[CLS]$ & Classification token \\
$c \in \mathbb{R}^{D_v}$ & $[CLS]$ token embedding \\
$s_i \in \mathbb{R}^{D_V}$ & Visual output embedding token \\

\midrule
\multicolumn{2}{l}{\textbf{Embeddings and Text Encodings}} \\
\midrule
$\phi_T(\cdot)$ & Text encoder \\
$[EOS]$ & End of sentence token \\
$T_E$ & Initial text embeddings \\
$L_T$ & Fixed context length \\
$D_T$ & Text embedding dimension \\
$t_{eos}$ &  End of sentence token \\
$t_{pad}$ & Padding token \\
$t_i \in \mathbb{R}^{D_T}$ & Text embedding token \\
\midrule
\multicolumn{2}{l}{\textbf{Projection Layers}} \\
\midrule
$W_V \in \mathbb{R}^{D_V \times D}$ & Trainable visual linear projection layer \\
$W_T \in \mathbb{R}^{D_T \times D}$ & Trainable test linear projection layer \\
$D$ & Projected dimension \\
$Z_V \in \mathbb{R}^{(1+N_V) \times D}$ & Projected visual tokens \\
$Z_T \in \mathbb{R}^{L_T \times D}$ & Projected text tokens \\
\midrule
\multicolumn{2}{l}{\textbf{Detector Transformer}} \\
\midrule
$\psi(\cdot)$ & Detector transformer \\
$t_h \in \mathbb{R}^D$ & Head token \\
$t_p \in \mathbb{R}^D $ & Target presence token \\
$c^{\prime}$ & Projected global visual tokens \\
$t^{\prime}_{eos}$ & Projected global text tokens \\
$E_{head}$ & Learnable head embeddings \\
$E_{presence}$ & Learnable presence embeddings \\
$s^{\prime} \in \mathbb{R}^{N_V\times D}$ & Projected visual patch tokens from $Z_V$ (excluding $c^\prime$) \\
$t^{\prime} \in \mathbb{R}^{(L_T - 2)\times D}$ & Projected text patch tokens from $Z_T$ (excluding $t^{\prime}_{eos}$ and padding) \\
$F\in \mathbb{R}^{(N_T+N_V+2) \times D}$ & Full input sequence F\\
$\psi(F)\in \mathbb{R}^{(N_T+N_V+2) \times D}$ & Output refined sequence of detector transformer \\

\midrule
\multicolumn{2}{l}{\textbf{Decoder}} \\
\midrule
$\hat{s} \in \mathbb{N_V \times D}$ & Refined visual patch tokens \\
$\hat{t_h} \in \mathbb{R}^D$ & Refined head tokens \\
x & normalized center x coordinate of head tracker \\
y & normalized center y coordinate of head tracker \\
w & normalized width of head tracker \\
h & normalized height of head tracker \\
$\hat{t_p} \in \mathbb{R}^D$ & Refined predict tokens \\
 
\midrule
\multicolumn{2}{l}{\textbf{Learning Objective}} \\
\midrule
$\mathcal{L}_{total}$ & Total loss \\
$\mathcal{L}_{gaze}$ & Gaze heatmap BCE loss \\
$\sigma$ & Standard deviation of 2D Gaussian \\
$\hat{Y}$ & Predicted heatmap \\
$N$ & Total number of pixels  \\
$p$ & Single pixel on the heatmap \\
$y_p$ & Ground-truth of p \\
$\hat{y_p}$ & Predicted values of p \\
$\mathcal{L}_{presence}$ & Target presence focal loss \\
$Y_{presence} \in {0, 1}$ & Ground truth target presence \\
$\hat{Y_{presence}} \in [0,1]$ & Predicted target presence \\
$\mathcal{L}_{focal}$ & Focal loss \\
$\mathcal{L}_{head}$ & Head box loss \\
$\mathcal{L}_{1}$ & Mean absolute error \\
$\mathcal{L}_{IoU}$ & GIoU loss \\
$b$ & Ground truth head box \\
$\hat{b}$ & Predicted truth head box \\
$\lambda_{l_1}$ & head object detection hyperparameter \\
$\lambda_{IoU}$ & head object detection hyperparameter \\
\midrule
\multicolumn{2}{l}{\textbf{Data Engine}} \\
\midrule
$x_{min}$ & x coordinate of top-left corner of the head box \\
$y_{min}$ & y coordinate of top-left corner of the head box \\
$x_{max}$ & x coordinate of bottom-right corner of the head box \\
$y_{max}$ & y coordinate of bottom-right corner of the head box \\
$g_x$ & x coordinate of ground truth gaze point \\
$g_y$ &  y coordinate of ground truth gaze point\\
    \hline
\end{longtable}

\end{document}